\documentclass[11pt, a4paper, logo, copyright, nonumbering]{alibaba}
\usepackage[authoryear, sort&compress, round]{natbib}
\usepackage{dblfloatfix}
\usepackage{ulem}
\usepackage{caption}
\usepackage{subcaption}
\usepackage{dramatist}
\usepackage{xspace}
\usepackage{pifont}
\usepackage{multirow}
\usepackage{tcolorbox}
\usepackage{xltabular}
\usepackage{longtable}
\usepackage{hyperref}
\usepackage{amsfonts}
\usepackage{amsmath}
\usepackage{amssymb}
\usepackage{lineno}
\usepackage{cleveref}
\usepackage{makecell}

\usepackage[bottom]{footmisc}

\usepackage{CJKutf8}
\usepackage{setspace}

\usepackage{dsfont}
\usepackage[table]{xcolor}
\usepackage{tabularx}
\usepackage{booktabs}
\newcolumntype{Y}{>{\centering\arraybackslash}X}

\makeatletter
\def\@BTrule[#1]{%
  \ifx\longtable\undefined
    \let\@BTswitch\@BTnormal
  \else\ifx\hline\LT@hline
    \nobreak
    \let\@BTswitch\@BLTrule
  \else
     \let\@BTswitch\@BTnormal
  \fi\fi
  \global\@thisrulewidth=#1\relax
  \ifnum\@thisruleclass=\tw@\vskip\@aboverulesep\else
  \ifnum\@lastruleclass=\z@\vskip\@aboverulesep\else
  \ifnum\@lastruleclass=\@ne\vskip\doublerulesep\fi\fi\fi
  \@BTswitch}
\makeatother

\addto\extrasenglish{
}

 {\begin{list}{}%
         {\setlength{\leftmargin}{#1}}%
         \item[]%
 }
 {\end{list}}
 
\reportnumber{001} 

\renewcommand{\today}{}

\newtcolorbox{examplebox}{
  colback=gray!10,
  colframe=gray!60,
  boxrule=0.8pt,
  arc=8pt,
  left=4pt, right=4pt, top=2pt, bottom=2pt
}

\title{Logics-STEM: Empowering LLM Reasoning via Failure-Driven Post-Training and Document Knowledge Enhancement}

\author[1*]{Mingyu Xu}
\author[1*]{Cheng Fang}
\author[1*]{Keyue Jiang}
\author[1,5*$\dagger$]{Yuqian Zheng}
\author[2,3]{Yanghua Xiao}
\author[2,4]{Baojian Zhou}
\author[1]{\authorcr Qifang Zhao}
\author[1]{Suhang Zheng}
\author[1]{Xiuwen Zhu}
\author[1,6$\dagger$]{Jiyang Tang}
\author[1]{Yongchi Zhao}
\author[1]{Yijia Luo}
\author[1]{Zhiqi Bai}
\author[1]{\authorcr Yuchi Xu}
\author[1]{Wenbo Su}
\author[1]{Wei Wang}
\author[1]{Bing Zhao}
\author[1]{Lin Qu}
\author[1]{Xiaoxiao Xu}

\affil[1]{Alibaba Group}
\affil[2]{Shanghai Key Laboratory of Data Science, Fudan University}
\affil[3]{College of Computer Science and Artificial Intelligence, Fudan University}
\affil[4]{School of Data Science, Fudan University}
\affil[5]{Georgia Institute of Technology}
\affil[6]{Nankai University}



\usepackage{amsmath,amsfonts,bm}
\usepackage{amssymb}

\def\eqref#1{equation~\ref{#1}}

\def\1{\bm{1}}

\DeclareMathAlphabet{\mathsfit}{\encodingdefault}{\sfdefault}{m}{sl}
\SetMathAlphabet{\mathsfit}{bold}{\encodingdefault}{\sfdefault}{bx}{n}

\def\gD{{\mathcal{D}}}

\def\gL{{\mathcal{L}}}

\def\gS{{\mathcal{S}}}

\renewcommand{\phi}{\varphi}

\renewcommand{\leq}{\leqslant}

\renewcommand{\epsilon}{\varepsilon}
\renewcommand{\imath}{\mathrm{i}}

\newlength{\restsubwidth}
\newlength{\restsubheight}
\newlength{\restsubmoreheight}
\newcommand{\rest}[2]{%
        \settowidth{\restsubwidth}{\ensuremath{#2}}
        \settoheight{\restsubheight}{\ensuremath{{}_{#2}}}
        \ensuremath{{#1\hskip 0.5pt}_{\vrule\kern2pt\parbox[b][%
        4pt][b]{\the\restsubwidth}{%
                        \ensuremath{{}_{#2}}}}}
        }

\newtheorem{theorem}{Theorem}[section]

\newtheorem{lemma}[theorem]{Lemma}

\begin{abstract}

We present Logics-STEM, a state-of-the-art reasoning model fine-tuned on Logics-STEM-SFT-Dataset, a high-quality and diverse dataset at 7.2M scale that represents one of the largest-scale open-source long chain-of-thought corpora.
Logics-STEM targets reasoning tasks in the domains of Science, Technology, Engineering, and Mathematics (STEM), and exhibits exceptional performance on STEM-related benchmarks with an average improvement of 4.68\% over the next-best model at 8B scale. We attribute the gains to our data-algorithm co-design engine, where they are jointly optimized to fit a gold-standard distribution behind reasoning. Data-wise, the Logics-STEM-SFT-Dataset is constructed from a meticulously designed data curation engine with 5 stages to ensure the quality, diversity, and scalability, including annotation, deduplication, decontamination, distillation, and stratified sampling. 
Algorithm-wise, our failure-driven post-training framework leverages targeted knowledge retrieval and data synthesis around model failure regions in the Supervised Fine-tuning (SFT) stage to effectively guide the second-stage SFT or the reinforcement learning (RL) for better fitting the target distribution.
The superior empirical performance of Logics-STEM reveals the vast potential of combining large-scale open-source data with carefully designed synthetic data, underscoring the critical role of data-algorithm co-design in enhancing reasoning capabilities through post-training.
We make both the Logics-STEM models (8B and 32B) and the Logics-STEM-SFT-Dataset-Open (excluding the private part, we release 5.3M and downsampled 1.6M versions) publicly available to support future research in the open-source community.
\end{abstract}

\addhflink{https://huggingface.co/Logics-MLLM/Logics-STEM-8B-SFT}{Model-SFT}
\addhflink{https://huggingface.co/Logics-MLLM/Logics-STEM-8B-RL}{Model-RL}
\addhflink{https://huggingface.co/datasets/Logics-MLLM/Logics-STEM-SFT-Dataset-Open-5.3M}{5.3M Dataset}
\addhflink{https://huggingface.co/datasets/Logics-MLLM/Logics-STEM-SFT-Dataset-Open-1.6M}{1.6M Dataset}

\begin{document}

\maketitle

\begingroup
\renewcommand\thefootnote{*}%
\footnotetext{Core Contributors.}%
\renewcommand\thefootnote{$^{\dagger}$}%
\footnotetext{Work done during internship at Alibaba Group.}%
\endgroup

\begin{figure}[htbp]    
  \centering            
  \includegraphics[width=\linewidth]{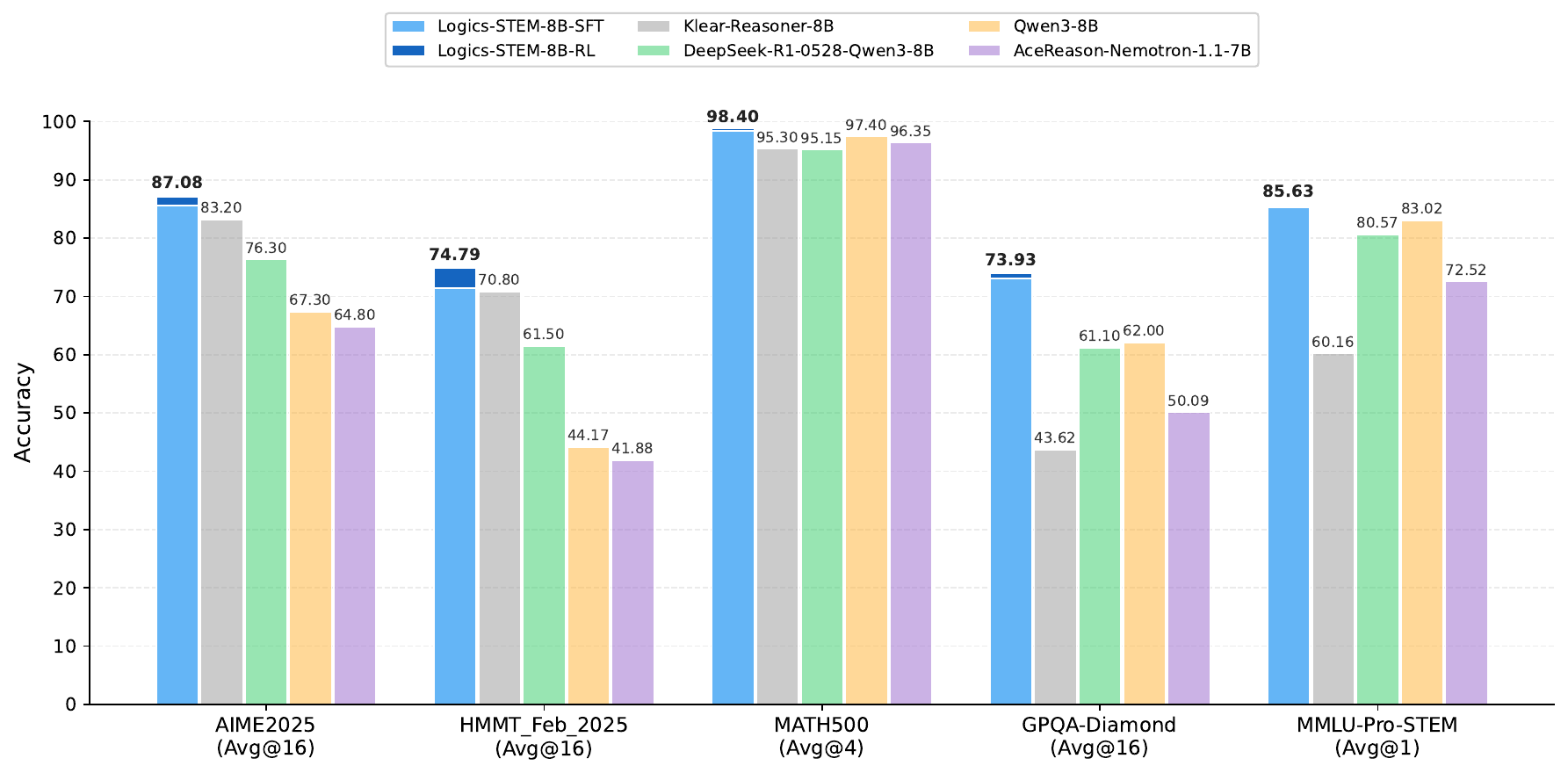} 
  \caption{\centering Model Performance Comparison on Representative STEM Benchmarks.}
  \label{fig:1}      
\end{figure}

\newpage

\section{Introduction}

In the past years, large language models (LLMs) like OpenAI’s o1 series \citep{jaech2024openai}, QwQ~\citep{qwq32b}, and DeepSeek-R1~\citep{guo2025deepseek} have demonstrated strong performance on challenging reasoning tasks in the field of mathematics and broader STEM. The reasoning capabilities in these models typically emerge from post-training techniques such as supervised fine-tuning (SFT) and/or reinforcement learning (RL) based on strong foundation models. However, while many models are open-sourced, the underlying post-training pipeline and training data curation remain undisclosed, creating both challenges and opportunities for future work. 

More recently, the open-source community has made substantial efforts to develop data-construction recipes and algorithmic strategies for cultivating advanced reasoning capabilities in small-scale models, leading to a series of notable works such as Klear-Reasoner~\citep{su2025klear}, Ring-Lite~\citep{DBLP:journals/corr/abs-2506-14731}, MiMo~\citep{coreteam2025mimounlockingreasoningpotential}, OpenThoughts~\citep{guha2025openthoughts}, Llama-Nemotron~\citep{bercovich2025llama}, and AceReason-Nemotron~\citet{liu2025acereason}, among others~\citep{fu2025areal, deepscaler2025, DBLP:journals/corr/abs-2503-10460}. However, despite these empirical successes, the community still lacks a unified framework to guide data curation and exploitation through post-training algorithms. It is widely recognized in the LLM community that ``data is the new oil'', and the algorithm can only succeed when it effectively captures the desired data distribution. This motivates a central question in training reasoning models:
\begin{center}
\textit{What does it take to build a Data-algorithm Co-design Engine for \\Reasoning Models in terms of Effectiveness, Efficiency, and Scalability?}
\end{center}
In this report, we address this question from both theoretical and engineering perspectives. We first provide a data-centric view of the widely adopted SFT-RL pipeline \citep{ouyang2022training} by framing it as a distribution-matching problem. We hypothesize that the first-stage SFT builds a strong proposal distribution that one draws samples for following usage; and the second stage post-training, no matter SFT or RL, shifts the model towards a gold-standard target distribution behind desirable properties such as reasoning ability. 

Building on this formulation, we push the boundaries of reasoning models by: (i) implementing a rigorous data curation pipeline to produce a scalable, broad-coverage, and high-quality long CoT dataset as a foundational proposal distribution; (ii) designing an optimized post-training pipeline that utilizes the curated data effectively and efficiently to improve the model’s reasoning capabilities.

Specifically, we curate reasoning data from publicly available datasets and further augment it with synthetic examples generated from documents parsed by Logics-Parsing~\citep{chen2025logicsparsingtechnicalreport}. 
Coupled with a fine-grained, difficulty-aware stratified sampling strategy, extensive experiments show that our curated Logics-STEM-SFT-Dataset already equips LLMs with strong foundational reasoning capabilities. Moreover, we adopt a failure-driven post-training paradigm to further align the model with the gold-standard reasoning distribution. Concretely, after the first-stage SFT, we perform targeted knowledge retrieval and synthesize data around the model’s failure regions to guide a second-stage SFT or RL. This yields two alternative pipelines: SFT–RL and SFT–SFT. We systematically test our method under both SFT-RL and SFT-SFT pipelines and show that our approach substantially improves the model’s reasoning ability.


Consequently, we present Logics-STEM, a reasoning model finetuned from Qwen3 that achieves outstanding performance on multiple key reasoning benchmarks. At 8B scale, as shown in \cref{fig:1}, Logics-STEM-8B outperforms Klear-Reasoner-8B, DeepSeek-R1-0528-Distill-8B and Qwen3-8B, scoring 90.42\% on AIME2024, 87.08\% on AIME2025, 74.79\% on HMMT2025, 62.5\% on BeyondAIME and 73.93\% on GPQA-Diamond.


In summary, the contributions of our work are listed as follows.

\begin{itemize}
\item We formulate the SFT-RL pipeline as a distribution matching problem and design a failure-driven post-training framework that leverages targeted knowledge retrieval and data synthesis around model failure regions to effectively and efficiently guide SFT and RL. 
    \item We design a data curation engine that effectively utilizes publicly available data and augments it with synthetic examples generated from documents. We then produce Logics-STEM-SFT-Dataset, a high-quality and diverse dataset that represents one of the largest‑scale open-source long chain-of-thought corpora at 10M scale.
    \item Our reasoning model, Logics-STEM, outperforms other open source models of comparable size on STEM reasoning benchmarks. We release Logics-STEM (8B and 32B) at both the SFT and RL stages, along with the open version of Logics-STEM-SFT-Dataset, to support further research and development within the open-source community.
\end{itemize}

\section{SFT-RL pipeline as a distribution matching}
\label{sec:dist_matching}

After pre-training, SFT followed by RL has become a widely adopted recipe for improving LLMs’ reasoning ability~\citep{ouyang2022training}. SFT is typically used to familiarize the model with long chain-of-thought (CoT) reasoning traces, while RL further aligns the model with human preferences or sharpens the policy distribution to produce more satisfactory responses with fewer samples~\citep{yue2025does, wen2025reinforcementlearningverifiablerewards}. From this perspective, the overall post-training procedure can be viewed as optimizing an expected objective estimated via Monte Carlo sampling. Let the training data consist of pairs $(x, y)$, where $x$ is the prompt and $y$ is the target output (e.g., a long-CoT response ending with a boxed final answer). Denote by $y^t$ the $t$-th token in $y$\footnote{We use subscripts $y_i$ to denote the $i$-th training example and superscripts $y^t$ to denote the $t$-th token within an example.}. We write the sample-level supervised loss as $\ell_\theta(x, y)$ (e.g., the negative log-likelihood, NLL), where $\theta$ are the model parameters. The training objective is then,
\begin{equation}
\label{eq:pop_risk}
\text{(Expected Risk)}\quad \gL^*(\theta)=\mathbb{E}_{(x,y)\sim P^*}\big[\ell_\theta(x,y)\big],
\end{equation}
where $P^*(x,y)$ denotes the (unknown) ideal target distribution over $(x,y)$. However, $P^*$ is rarely directly accessible in practice, as it corresponds to a ``perfect'' or gold-standard distribution for solving reasoning tasks. Instead, one only observes a surrogate dataset $\gD$, which is drawn from some distribution $P_0(x,y)$ with $P_0 \neq P^*$. This dataset is typically assembled from diverse sources without a unified curation criterion, as we elaborate in~\cref{sec: methodology}. Consequently, $P_0$ can differ substantially from $P^*$, inducing a distribution mismatch between the available training data and the ideal target distribution that would best support reasoning.

\paragraph{The mismatch between target and training distribution.} Practically, a biased optimization is done through minimizing $\mathbb{E}_{(x,y)\sim P_0}\big[\ell_\theta(x,y)\big]$ via Monte Carlo estimation, for which one samples mini-batches $\{x_i, y_i\}_{i=1}^B$ from $P_0$ and approximate the expectation by the empirical mean. To eliminate the bias between target and training, we first look at the issues that cause high expected risk. We consider an importance sampling formula and reformulate~\cref {eq:pop_risk} as
\begin{equation}
\label{eq:iw_grad_form}
\gL^*(\theta)
=\mathbb{E}_{(x,y)\sim P^*}\big[\ell_\theta(x,y)\big]=
\mathbb{E}_{(x,y)\sim P_0}\!\left[
\underbrace{\frac{P^*(x,y)}{P_0(x,y)}}_{\text{Density Ratio}}\underbrace{\,\ell_\theta(x,y)
}_{\text{Sample-wise Loss}}\right],
\end{equation}
The high risk of $\gL^*$ is caused by two issues: 1) high density ratio $\frac{P^*(x,y)}{P_0(x,y)}$, which means the region is underestimated or rarely seen in training but important in the target distribution; and 2) high sample-wise loss $\ell_\theta(x,y)$, which means the model completely fails in these samples. 

\begin{examplebox}
\textbf{Remark: } From the formula above, we hypothesize that: 1) the first stage SFT is trying to fit the model to a good proposal distribution $P_0$, and 2) the second stage RL is trying to explore the region that has a high density ratio and shifts the distribution towards the golden distribution. 
\end{examplebox}

We elaborate on the remark a bit. SFT fits the model to $P_0$ through a large-scale dataset $\gD \sim P_0$, which provides good proposal distribution with broad coverage and grants the model with general reasoning abilities. However, at this stage, the mismatch between $P_0$ and $P^*$ would make the optimization biased, thus performing suboptimally on specific tasks. The second stage training is used to implicitly eliminate the bias. For instance, a vanilla policy gradient can be considered as fitting a distribution depending on the advantage function $A(x,y)$~\citep{DBLP:conf/icml/0011C00LZ23},
\begin{equation}
P^\prime(y \mid x) \propto P_0(y \mid x) e^{\beta A(x,y)},
\end{equation}
where $\beta$ is the regularization weight. With a proper selection, $P^\prime$ can be considered as a good surrogate for the target distribution of $P^*$, and the second-stage RL is used to shift the distribution towards $P^*$. We provide a detailed discussion in Appendix~\ref{apdx: ext_RL}. 

With this understanding, the second-stage RL can even be replaced by SFT with proper algorithm design, as we will illustrate theoretically in~\cref{sec:failure_sampling} and empirically in~\cref{sec: experiment}. As such, we propose two key principles for building a data-algorithm co-design engine for strong reasoning models:

(1) \textbf{Stage-1 SFT should produce a strong proposal distribution.} To this end, we empirically design a data engine that generates high-quality training data, guided by practical experience and engineering insights. \Cref{sec: methodology} details this data engine.

(2) \textbf{Stage-2 post-training should efficiently shift the model toward the target distribution.} In~\cref{sec:failure_sampling}, we propose a more effective and efficient way to leverage data to shift the model toward $P^*$. Specifically, we introduce a failure-driven resampling for second-stage post-training, enabling target-oriented optimization to enhance reasoning ability.

\section{Reasoning Data Engine}
\label{sec: methodology}

\begin{figure}[htbp]
  \centering   \includegraphics[width=1\linewidth]{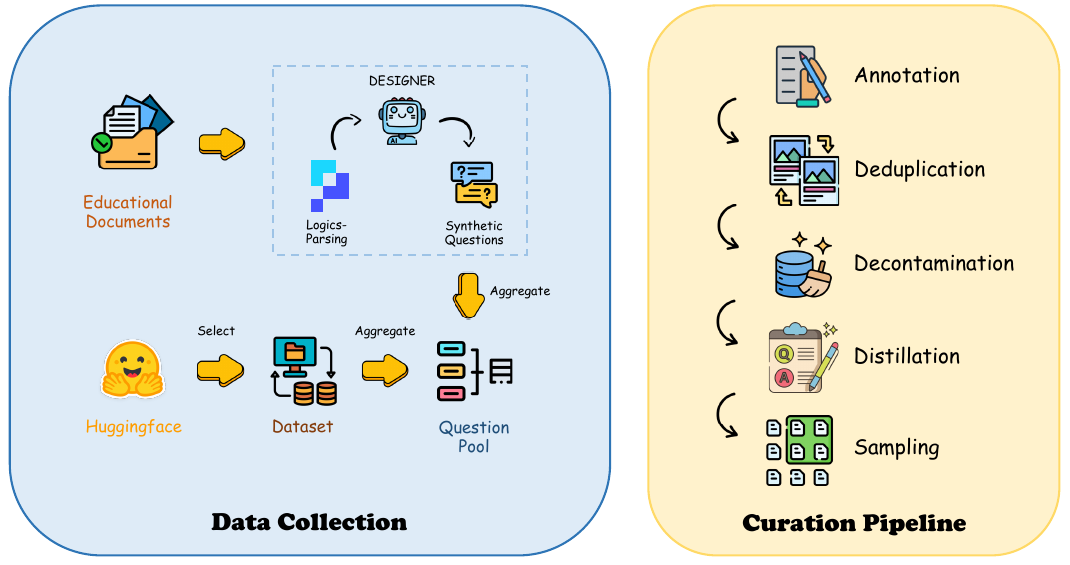} 
  
  \caption{\centering Overview of Long CoT Data Engine}
  \label{fig:stage_1_engine} 
\end{figure}

The central goal of the first-stage SFT is to build a diverse, effective, and scalable data engine that learns a proposal distribution $P_0$ tailored to reasoning tasks. To this end, we construct long CoT datasets that are rich in reasoning content and generalize well across a wide range of reasoning benchmarks to ensure the $P_0$ has a broad support. As shown in~\cref{fig:stage_1_engine}, our pipeline collects questions from selected sources and produces high-quality reasoning question-response pairs through a series of curation steps.

\subsection{Data Curation Pipeline}

\paragraph{Data Collection.}Previous studies~\citep{guha2025openthoughts,liu2025acereason,openr1} have demonstrated the effectiveness of open-source data. To fully leverage open-source data and ensure diversity, we aggregate questions from a wide range of highly regarded and frequently cited sources on Huggingface\footnote{https://huggingface.co/}, as listed in Appendix~\ref{a:source}. For the purpose of preventing potential quality risks, all early synthetic subsets and multi-modal subsets are excluded. Additional, we utilize DEGISNER~\citep{liu2025designer} to incorporate synthetic datasets from books and webpages (DLR-Book and DLR-Web in~\citet{liu2025designer}) to further scale dataset volume and enhance diversity. All PDF-format documents are parsed into plain text by Logics-Parsing~\citep{chen2025logicsparsingtechnicalreport} and subsequently processed through the DESIGNER synthesis pipeline. 

\paragraph{Annotation.} To unify data curation for the SFT and to further accommodate the subsequent RL with verifiable reward (RLVR) stages, we prompt Qwen3-235B-Instruct-2507\footnote{https://huggingface.co/Qwen/Qwen3-235B-A22B-Instruct-2507} to annotate each sample (including question, meta-data, solution, and answer) across multiple dimensions: (1) whether the question is valid and unambiguous, (2) the discipline and domain associated with the question, (3) the educational level of the question, (4) the answer type,  and (5) the verifiable answer (if exists). 
Samples that do not pass the validity and unambiguity check are subsequently filtered out.
Details of annotation are provided in Appendix~\ref{a:annotation}.

\paragraph{Deduplication.}
To ensure diversity, deduplication is performed at multiple granularities, including both exact and near-duplicate removal. Firstly, we generate an MD5 fingerprint for each question and eliminate redundant samples with identical fingerprints across all sources. Moreover, we apply MinHash-based deduplication using 24 bands with a bandwidth of 10 to identify and eliminate near-duplicate samples. Specifically, within each group of samples sharing the same MinHash bucket, samples with valid questions and verifiable answers are prioritized for retention. Ultimately, approximately two-thirds of the data (about 10 million instances) are retained.

\paragraph{Decontamination.}
We perform decontamination against the evaluation benchmarks to eliminate potential contamination in the training data, employing both MinHash-based and N‑gram based methods. Any training sample sharing the same MinHash bucket or an identical 13‑gram with evaluation samples is removed.

\paragraph{Response Distillation.}
To balance computational efficiency and response accuracy, Qwen3-235B-A22B-Thinking-2507\footnote{https://huggingface.co/Qwen/Qwen3-235B-A22B-Thinking-2507} serves as the teacher model to distill reasoning responses for each question. 
Generation configurations are detailed in Appendix \ref{a:generation}. To suppress systematic bias during subsequent training, we discard responses that exhibit degenerative repetition. Concretely, any response with an n-gram duplication ratio exceeding a predefined threshold is excluded from the training dataset. Optional further answer verification is conducted using \texttt{math-verify} library\footnote{https://github.com/huggingface/Math-Verify}. In cases where responses do not match the standard verifiable answers, the teacher model is employed to regenerate the response once more. The strategy of filtering out incorrect responses is not adopted, as empirical results indicate that it leads to degraded performance, consistent with the conclusion of OpenThoughts~\citep{guha2025openthoughts}.

\subsection{Weighted Stratified Sampling}
\label{sec:stratified sampling}
Despite the implementation of extensive filtering strategies, the volume of data generated by our pipeline remains a significant challenge for model training, thereby necessitating further data sampling. Inspired by OpenThoughts~\citep{guha2025openthoughts}, we employ a \textbf{difficulty-based weighted stratified sampling strategy} on STEM-related data to achieve a balance between reasoning density and data diversity. Specifically, token length of the response is adopted as a natural and effective proxy for question difficulty. Our subsequent experiment corroborates the finding of OpenThoughts that length-based sampling outperforms annotation-based alternatives. 
\\

However, we further observe that purely length-based sampling leads to degraded model performance on relatively elementary benchmarks (See Section \ref{sec: sampling strsategy}). To address this issue, we adopt a stratified sampling strategy for data mixing. In practice, we compute the quantiles of response token lengths and retain all samples above the 75th percentile. Samples between the 50th and 75th percentiles are downsampled by 50\%, while those between the 20th and 50th percentiles are downsampled by 10\%. This approach ensures a high reasoning density within the training set while maintaining overall data diversity.

\subsection{The curated Dataset Statistics}

\begin{figure}
    \centering
    \includegraphics[width=1\textwidth]{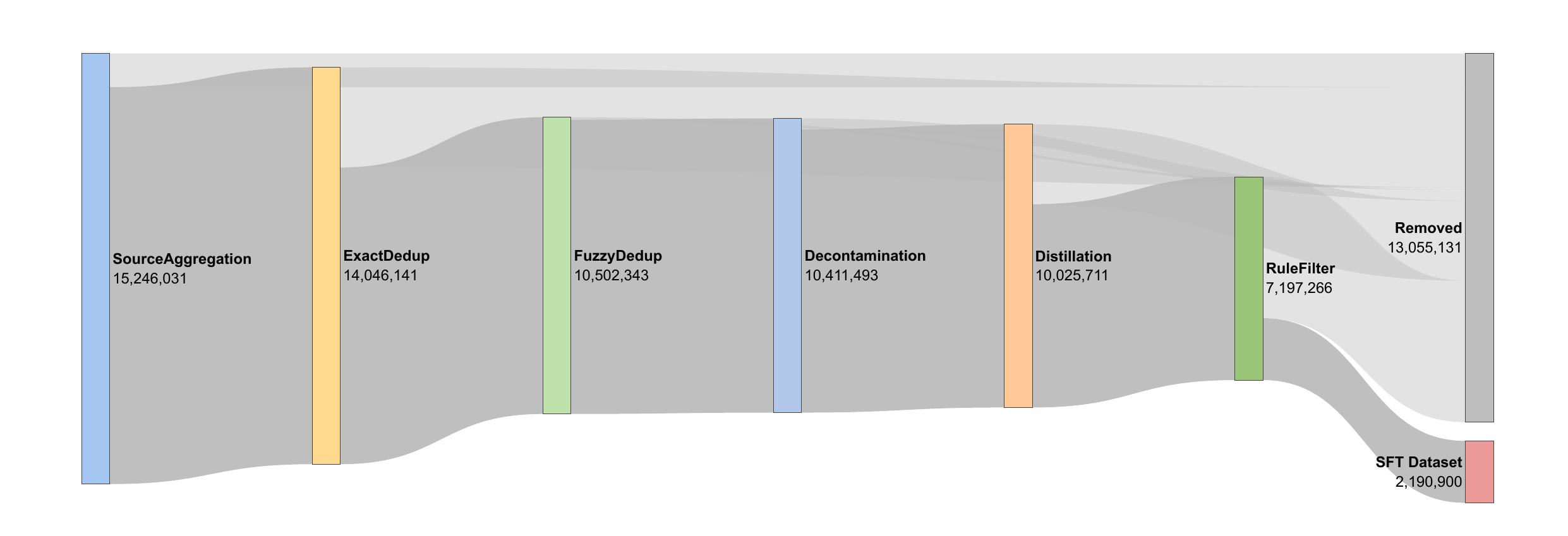}
    \caption{The dataflow of Logics-STEM-SFT-Dataset: From original sources to ready-to-use.}
    \label{fig: construction of Logics-STEM}
\end{figure}

\begin{figure}
    \centering
    \begin{subfigure}[b]{0.59\textwidth}
        \centering
        \includegraphics[width=\linewidth]{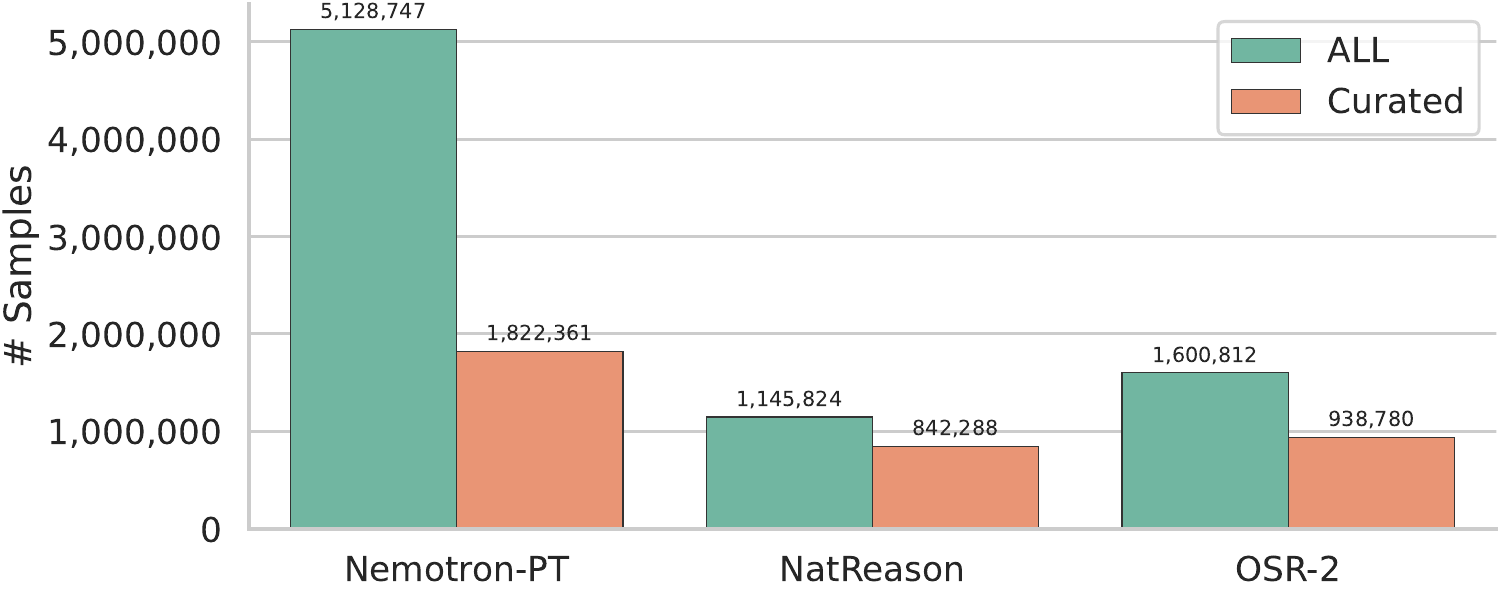}
        \caption{\centering The dataset statistics after curation.} 
        \label{fig:counts}
    \end{subfigure}
    \hfill
    \begin{subfigure}[b]{0.40\textwidth}
        \centering
        \includegraphics[width=\linewidth]{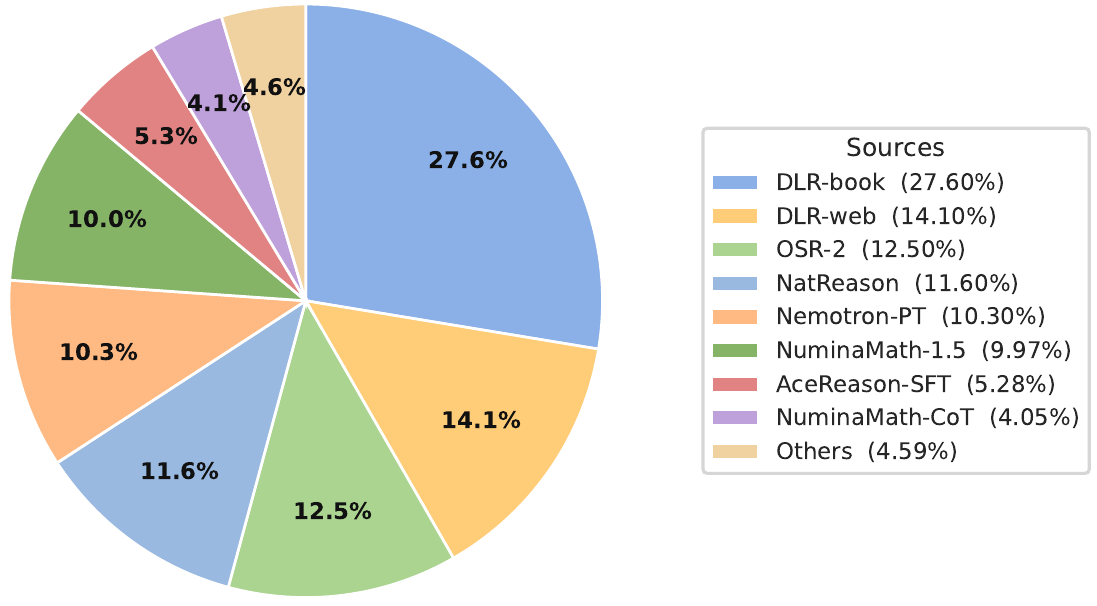}
        \caption{\centering The source datasets distribution} 
        \label{fig:pie}
    \end{subfigure}
    \caption{Comparison of source dataset distribution and curation statistics. Nemotron represents for nvidia-Llama-Nemotron-Post-Training-Dataset~\citep{bercovich2025llama}, NatReason for naturalReasoning~\citep{yuan2025naturalreasoning}, OSR-2 for OpenScienceReasoning-2.}
    \label{fig:comparison}
\end{figure}

After these fine-grained steps, we obtain \textit{Logics-STEM-SFT-Dataset-7.2M} before stratified sampling. To the best of our knowledge, it is among the largest open-source long chain-of-thought corpora that is high-quality, diverse, and most importantly, ready for direct use. The overall data flow is shown in~\cref{fig: construction of Logics-STEM}. We further report the source dataset distribution in~\cref{fig:pie}, and summarize dataset statistics before and after curation for the large-scale open-source sources in~\cref{fig:counts}, highlighting our curation engine’s emphasis on both data quality and diversity. After stratified sampling, we obtain \textit{Logics-STEM-SFT-Dataset-2.2M}, comprising 1.05M math samples and 1.14M broader-STEM samples distilled from a teacher model to preserve strong reasoning traces.

\section{Failure-Driven Post-training via Knowledge Enhancement}
\label{sec:failure_sampling}

The widely used post-training pipeline~\citep{ouyang2022training} consists of two stages: SFT followed by second-stage RL. From our distribution-matching perspective (\cref{sec:dist_matching}), the second stage aims to better approximate the target distribution, and can be implemented via either SFT or RL. Accordingly, in this section we introduce a failure-driven post-training pipeline that is both effective and efficient. Our paradigm is theoretically grounded to better fit the gold-standard distribution for reasoning ability, and is sample-efficient as it correctly leans towards the regions that contribute most to distribution matching.

\subsection{Algorithmic Design}

\subsubsection{First-stage Supervised Fine-tuning} 

Once the data are ready to use, we conduct the first-stage SFT. Considering the token-level decomposition and the supervised loss as the negative log-likelihood (NLL), the training will yield an optimized parameter $\theta_1$,
\begin{equation}
\label{eq:stage1-sft}
\theta_1=\arg\min_{\theta}\;\mathbb{E}_{(x,y)\sim P_0}\big[\ell_\theta(x,y)\big],
\quad
\ell_\theta(x,y)=-\sum_{t=1}^{T}\log \pi_\theta\!\left(y^t\mid x,y^{<t}\right).
\end{equation}
where $T$ denotes the sequence length, $\pi_\theta(y^t|x, y^{<t})$ denotes the probability of generating the target token $y^t$ under current context $(x,y^{<t})$ (policy in RL).
This would fit the model towards the proposal distribution $P_0$.

\subsubsection{The Failure-driven Second-stage Training}

Although SFT improves LLM performance by fitting the model to a proposal distribution $P_0$, it remains limited on complex reasoning tasks, especially high-difficulty, cross-domain STEM problems where models often fail to produce correct answers. This problem reveals that there is still a distance between $P_0$ and $P^*$. Thus, to solve the mismatch between $P_0$ and $P^*$, we conduct a second-stage post-training via the following steps: evaluation $\rightarrow$ knowledge retrieval $\rightarrow$ data synthesis $\rightarrow$ continued training. The evaluation helps us identify the regions $(x,y)$ that cause high overall loss (failure regions). Once the failure region is found, we retrieve from external documents for knowledge enhancement and synthesize more query-response pairs to provide more samples and compensate for the density ratio in~\cref{eq:iw_grad_form}. Then, a second-stage training is done to minimize $\ell_\theta(x,y)$ in that region.

\paragraph{Evaluation for failure region identification.} After the first-stage SFT and obtaining $\theta_1$, the model has already gained the basic information. To find the regions (i.e., query-response pairs) that the model underestimated (high density ratio between $P^*$ and $P_0$) or highly biased (high sample-wise loss $\ell_\theta(x,y)$), we evaluate the model on a distribution constructed from gold-standard evaluation tasks (such as AIME2025~\citep{AIME2025}, MMLU~\citep{wang2024mmlu} etc.) denoted as a distribution $Q$, and collect failure cases from this evaluation distribution. We define a failure function $w_{\theta_1}(x)\in\mathbb{R}_+$ to evaluate the value of the current sample to the model training, which in our implementation is a binary indicator of verified incorrectness:
\begin{equation}
\label{eq:binary-failure}
w_{\theta_1}(x)\in\{0,1\}, \quad 
w_{\theta_1}(x)=1 \;\;\text{iff the verified final answer of $\pi_{\theta_1}$ on $x$ is incorrect}.
\end{equation}
This induces a failure-driven query distribution
\begin{equation}
\label{eq:failure-biased-q}
Q_{\theta_1}(x)=\frac{Q(x)\,w_{\theta_1}(x)}{\mathbb{E}_{x\sim Q}[w_{\theta_1}(x)]},
\end{equation}
which increases the sampling probability of regions where the current model performs poorly. Under the assumption that $Q$ is a proper surrogate for $P^*$, we can show that the new query distribution can better minimize the expected risk in~\cref{eq:iw_grad_form}. We left the proof in Appendix~\ref{apdx: RL_theory}.

\paragraph{Retrieval kernel induced by embeddings.}
Failure resampling alone does not explicitly provide missing knowledge, as it cannot expand the support of $P_0$, and thus can only provide a better approximation to the existing set. We therefore introduce an external document corpus $\mathcal{K}=\{d\}$ to enhance the second-stage training, where $d$ denotes the external knowledge document instances. This external document corpus can be retrieved to further scale up the training dataset. 

We formalize retrieval as sampling via a kernel.
Let $\phi(\cdot)$ be an embedding model and define similarity (via cosine similarity) as 
$s(x,d)=\langle \phi(x),\phi(d)\rangle$, we can define the conditional sampling distribution over external documents in the knowledge base as
\begin{equation}
\label{eq:kernel-full}
K_{\tau}(d\mid x)=\frac{\exp\!\big(s(x,d)/\tau\big)}{\sum_{d'\in \mathcal{K}}\exp\!\big(s(x,d')/\tau\big)},
\end{equation}
where $\tau$ is the temperature to adjust the distribution. 
In practice, to enhance the sparsity of retrieval to improve efficiency, we use a top-$k$ truncated variant looks like:
\begin{equation}
\label{eq:kernel-topk}
K_{\tau,k}(d\mid x)\propto \exp\!\big(s(x,d)/\tau\big)\cdot \mathbf{1}\!\left[d\in \mathrm{Top-}k(x)\right].
\end{equation}
Intuitively, $K_{\tau,k}(d\mid x)$ specifies how likely a failure case $x$ ``activates'' a related knowledge document $d$, thereby transferring sampling mass from failure regions to their neighborhoods in the knowledge space.

\paragraph{Data synthesis and the induced synthetic training distribution.}
Given a retrieved document instance $d$, we synthesize a new training example $(x',y')$ via powerful LLMs such as DeepSeek R1~\citep{guo2025deepseek}. We view synthesis as another conditional kernel $G((x^\prime, y^\prime)\mid d)$ and apply an acceptance filter $a(x^\prime, y^\prime)\in\{0,1\}$ (e.g., answer-consistency verification), yielding
$G_{\mathrm{acc}}(x^\prime, y^\prime\mid d)\propto G(x^\prime, y^\prime\mid d)\,a(x^\prime, y^\prime)$.

\paragraph{Second Stage Post-training.}Combining failure sampling, retrieval, and synthesis, we obtain the synthetic data distribution:
\begin{equation}
\label{eq:psyn}
P_{\mathrm{syn}}(x^\prime, y^\prime)
=
\mathbb{E}_{x\sim Q_{\theta_1}}\;
\mathbb{E}_{d\sim K_{\tau,k}(\cdot\mid x)}\!\left[
G_{\mathrm{acc}}(x^\prime, y^\prime\mid d)
\right],
\end{equation}
or equivalently,
\begin{equation}
\label{eq:psyn-sum}
P_{\mathrm{syn}}(x^\prime, y^\prime)
=
\sum_x Q_{\theta_1}(x)\sum_d K_{\tau,k}(d\mid x)\,G_{\mathrm{acc}}(x^\prime, y^\prime\mid d).
\end{equation}
Thus, the second stage trains on a distribution that mixes $P_\text{syn}$ and $P_0$ with percentage $\lambda$. Rather than relying on implicit distribution matching via RL, we argue that SFT can be used for explicit distribution matching in the second stage, yielding the following training objective:
\begin{equation}
    \gL_1(\theta) \triangleq \mathbb{E}_{(x,y)\sim P_1}\big[\ell_\theta(x,y)\big],\quad P_1 = \lambda P_0 + (1-\lambda) P_{\mathrm{syn}}.
\end{equation}
Equivalently, this is done via sampling datapoints from $P_0$ and $P_\text{syn}$ through a percentage of $\lambda$. We prove the following theorem to show that, under mild conditions, the failure-driven post-training yields better expected loss. 

\begin{examplebox}
\begin{theorem}[(Informal) Failure-driven training minimizes the expected loss.]
\label{cor:failure_alignment}
Let $\nabla \gL^*(\theta)$ be the ideal target graident, $\nabla \gL_{1}(\theta)$ be the stage-2 gradient under the synthetic
distribution $P_{1}$, and $\nabla \gL_0(\theta)$ be the gradient under the original
distribution $P_0$. Assuming the gradient is normalized such that $\|\nabla \gL_1(\theta)\|=\|\nabla \gL_0(\theta)\|$ and $Q\approx P*$, the failure-driven construction of $P_1$ makes the gradients better aligned with the target gradient, i.e.,
\begin{equation}
\langle \nabla \gL^*(\theta), \nabla\gL_1(\theta)\rangle
\ge
\langle \nabla \gL^*(\theta), \nabla \gL_0(\theta)\rangle,
\end{equation}
Then, for sufficiently small learning rate $\eta$, one stage-2 optimization step using $P_{1}$ yields no larger target risk than using $P_0$:
\begin{equation}
\gL^*(\theta-\eta \nabla  \gL_1(\theta)) \le \gL^*(\theta-\eta \nabla \gL_0(\theta)).
\end{equation}
\end{theorem}
\end{examplebox}
We left the proof and detailed derivation in~\cref{app:grad_alignment}.

\subsubsection{Extension to Reinforcement learning}

Our framework is derived under the purpose of dealing the mismatch between training and target distribution, thus it can be naturally extended to design a second stage training under reinforcement learning via verified reward (RLVR). We do not elaborate the theoretical proof in the main text, as it follows a similar structure as failure-driven SFT, but only show the algorithm here. The detailed derivation can be found in~\cref{apdx: ext_RL}

\paragraph{Stage-2 RLVR.} We first sample the prompts from the marginal distribution $P_1(x)$.
Then, for a prompt $x$, the model samples an output $y\sim \pi_\theta(\cdot\mid x)$ and receives a verified reward as the advantage function $A$, where
\begin{equation}
\label{eq:rl-binary}
A(x,y)\in\{0,1\},
\quad
A(x,y)=1 \;\;\text{iff the final answer in $y$ is verified correct}.
\end{equation}
The vanilla RL objective is
\begin{equation}
\label{eq:stage2-rl}
\gL(\theta) = \mathbb{E}_{x\sim P_{1}(x)}\;
\mathbb{E}_{y\sim \pi_\theta(\cdot\mid x)}\big[A(x,y)\big].
\end{equation}

Practically, we test our model using GRPO~\citep{shao2024deepseekmathpushinglimitsmathematical} and DAPO~\citep{yu2025dapoopensourcellmreinforcement}, which we defer the introduction in~\cref{sec:RL_imp}.

\subsection{Engineering Implementation}
\label{sec: engineering_imple}

\begin{figure}[htbp]    
  \centering            
  \includegraphics[width=1\linewidth]{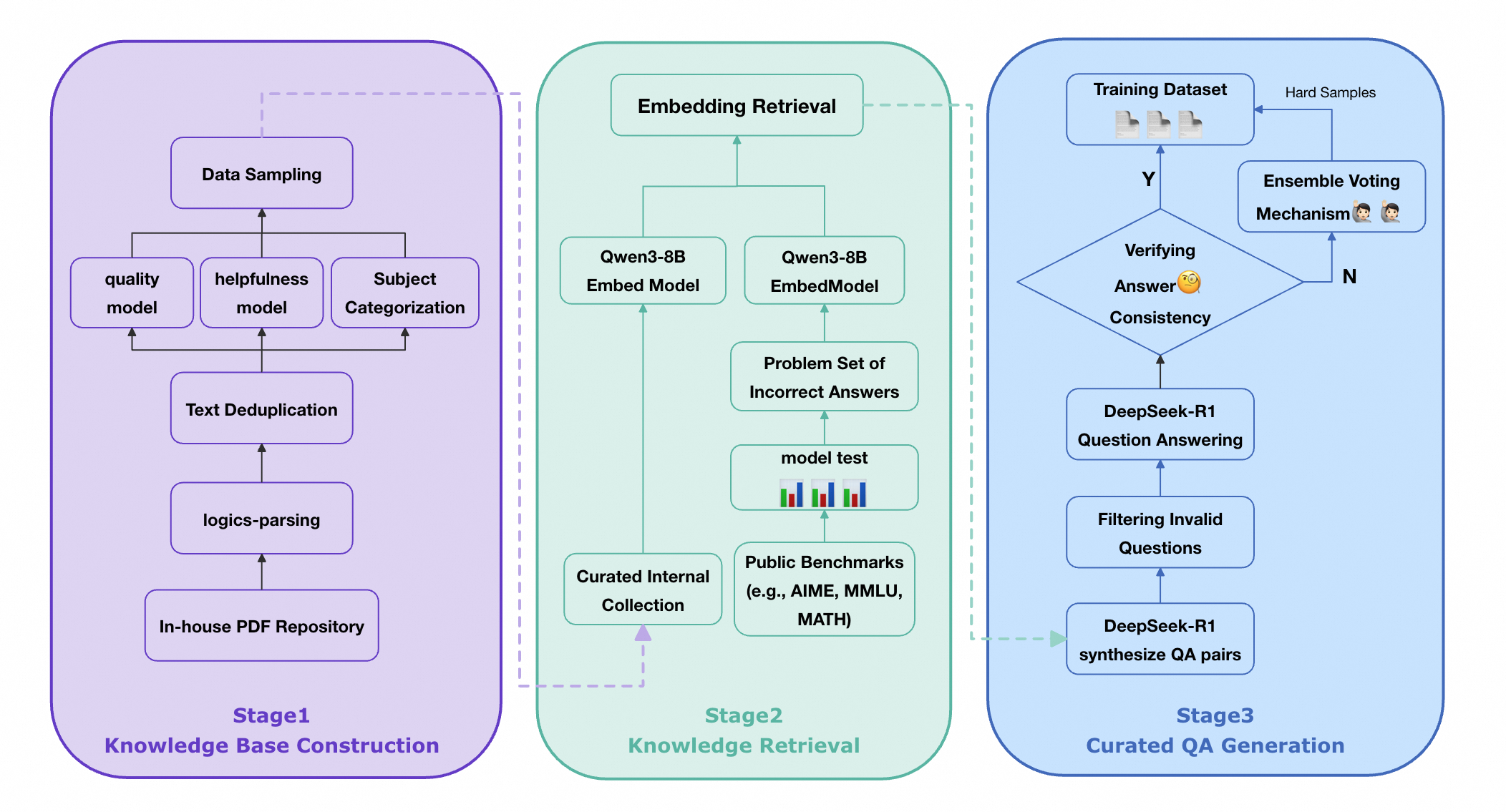} 
  \caption{\centering Overview of Knowledge-Driven Data Engine}
  \label{fig:rlvr_dataset}      
\end{figure}

We then introduce our engineering implementation. 
The central tenet of second-stage training and data usage is to identify the underestimated region from the failure cases and remediate them using external, high-fidelity knowledge sources. This is done through three steps: 1) knowledge base construction, 2) knowledge retrieval engine, and 3) curated question-answer (QA) pair generation. The overall pipeline is illustrated in~\cref{fig:rlvr_dataset}. Notably, data employed in the second stage is generated automatically without any human annotation.

\subsubsection{Knowledge Base Construction}
We curated an internal multi-source PDF corpus encompassing academic papers, technical monographs, authoritative textbooks, and technical reports. To preserve both structural fidelity and semantic integrity, we employed LogicsParsing(\cite{chen2025logicsparsingtechnicalreport}),a proprietary, high-precision PDF parsing tool, to convert raw documents into structured HTML format. Subsequently, we applied a multi-dimensional quality filtering mechanism:
\begin{itemize}
    \item A text quality model evaluated linguistic fluency;
    \item A usefulness model quantified information richness;
    \item A fine-grained subject classification model assigned domain-specific labels.
\end{itemize}
Only documents exhibiting high fluency, high knowledge density, and unambiguous domain specificity were retained, culminating in a high-quality, structurally coherent knowledge base.

\subsubsection{Knowledge Retrieval}
We first evaluate the SFT model on multiple established STEM benchmarks, including AIME2025 \citep{AIME2025}, MATH500, MMLU~\citep{wang2024mmlu} that works as the gold-standard distribution $Q$. We then collect questions answered incorrectly as ``knowledge deficiency'' samples. These erroneous instances are then encoded into dense vector representations using the Qwen3-8B-Embed\footnote{https://huggingface.co/Qwen/Qwen3-Embedding-8B}. Concurrently, every document in the knowledge base is embedded using the same model. Leveraging a vector retrieval algorithm, we retrieve the top-30 semantically most relevant documents for each incorrect question, ensuring high topical, conceptual, or methodological alignment between the query and retrieved content.

\subsubsection{Data Synthesis}
Utilizing the DeepSeek-R1~\citep{guo2025deepseek} as the synthesis kernel $G((x^\prime, y^\prime)\mid d)$, we automatically generate two Query-Response pairs per retrieved document, each requiring deep reasoning based on the document's core knowledge points and logical structure. To guarantee answer accuracy and consistency, a dual-verification mechanism is implemented to serve as the acceptance filter:
\begin{itemize}
    \item In the first pass, a question and a preliminary answer are generated jointly;
    \item In the second pass, the model is prompted with the question alone to generate an independent answer;
    \item Only samples yielding identical answers across both passes are retained as high-confidence training instances.
\end{itemize}
For inconsistent cases, we aggregate additional model responses and applied a majority-voting mechanism to determine a consensus answer; These samples are incorporated into the training set as “hard examples.” Furthermore, the length of the Chain-of-Thought (CoT) reasoning trace generate during the second answer synthesis serves as a proxy indicator for difficulty, as longer CoT sequences typically signify more complex reasoning steps and greater challenge. This criterion facilitates the selection of high-difficulty samples, ensuring the training data not only possesses high reliability but also effectively stimulates the model's potential for advanced reasoning.
Through this rigorous pipeline, we ultimately synthesize a dataset comprising approximately 30K high-quality, high-difficulty, knowledge-aligned QA pairs. This dataset is subsequently used to train the SFT model, resulting in substantial improvements in robustness and generalization on complex reasoning tasks. 


\subsubsection{Reinforcement Learning Implementation}
\label{sec:RL_imp}

We test GRPO~\citep{shao2024deepseekmathpushinglimitsmathematical} and DAPO~\citep{yu2025dapo} for our framework, which defines the policy ratio $r_{i, t}$ and the renormalized advantage function $\hat{A}_{i}$ as, 
\begin{equation}
r_{i, t}(\theta)=\frac{\pi_\theta\left(y_i^t \mid x, y_i^{<t}\right)}{\pi_{\theta_{\text {old }}}\left(y_{i}^t \mid x, y_{i}^{<t}\right)}, \quad \hat{A}_{i}=\frac{A_i-\operatorname{mean}\left(\left\{A_i\right\}_{i=1}^G\right)}{\operatorname{std}\left(\left\{A_i\right\}_{i=1}^G\right)} .
\end{equation}
We adopt the clip-higher strategy alongside batch-level reward normalization. Specifically, for each $x\sim P_1(x)$, we sample multiple $\{y_i\}_{i=1}^G$, and defines the 
\begin{equation}
\label{eq: DAPO}
    R(x,y) =\left[\frac{1}{\Sigma_{i=1}^{G}|y_i|}\sum_{i=1}^{G}\sum_{t=1}^{|y_i|}\min(r_{i, t}(\theta)\hat{A}_{i}, \text{clip}(1+\epsilon, 1-\epsilon, r_{i, t}(\theta))\hat{A}_{i})\right]
\end{equation}


In addition to the answer correctness reward as in~\cref{eq:rl-binary}, we add format compliance and reasoning-length reward to the RL.

\paragraph{Answer Correctness.} We use the open-source framework ROLL~\citep{wang2025reinforcementlearningoptimizationlargescale} to compute verified rewards. For math problems, we use \texttt{math-verify} to compare model outputs against ground-truth answers and assign a binary reward as advantage $A_i\in\{0,1\}$. For multiple-choice questions, we extract the option in $\texttt{\textbackslash boxed}\{...\}$ and match it to the ground truth. Following \citet{yu2025dapoopensourcellmreinforcement}, we apply dynamic sampling to filter problematic training instances by (i) dropping over-length responses and (ii) excluding prompts whose response-group mean scores fall outside a preset range. Hyperparameters are summarized in Table~\ref{tab: hyperparam for rlvr}.

\paragraph{Length and Repetition Aware Reward.} We observe that even when the final answer is incorrect, longer and more explicit CoT traces often facilitate self-correction and gradual convergence to the correct solution. Accordingly, we encourage deeper reasoning by assigning a length-based reward: for incorrect answers, the reward increases monotonically with the CoT length; for correct answers, we grant the maximum length reward by default to avoid discouraging concise yet correct reasoning. We also find that CoT generations can be redundant. To improve diversity and information density, we incorporate an $n$-gram–based repetition penalty into the reward.

Formally, let $A\in\{0,1\}$ indicate answer correctness, $\ell$ be the reasoning-trace length, and $\ell_{\min},\ell_{\max}$ be length bounds. We define the normalized length score
\[
s(T,A)=
\begin{cases}
1, & A=1 \ \text{or}\ T\ge T_{\max},\\
0, & A=0 \ \text{and}\ T\le T_{\min},\\
\dfrac{T-T_{\min}}{T_{\max}-T_{\min}}, & \text{otherwise}.
\end{cases}
\]
The final length-aware reward is $r_{\text{len}} = s(T,A)\cdot \rho,$
where $\rho\in(0,1]$ is a penalty factor computed from the $n$-gram repetition rate of the generated text.

\section{Experiments}
\label{sec: experiment}

With systematic experiments, we show that our framework not only yields significant gains in reasoning accuracy on STEM benchmarks but also establishes a novel, reproducible, transferable, and highly efficient paradigm for data construction in the post-training phase.

\subsection{General Evaluation}
\label{sec:eval}

\subsubsection{Benchmarks}
\label{sec:benchmark}

Models are evaluated on representative reasoning benchmarks spanning mathematics and broader STEM disciplines. For mathematical reasoning, we assess performance on AIME2024 \citep{AIME2024}, AIME2025\citep{AIME2025}, HMMT\_Feb\_2025, BRUMO2025, BeyondAIME\citep{beyondaime}, and MATH-500. For remaining STEM domains, we adopt GPQA-Diamond \citep{rein2024gpqa}, R-Bench \citep{rbench}, MMLU-Pro \citep{wang2024mmlu}, and CMMLU \citep{li2023cmmlu} as benchmarks. Particularly for MMLU-Pro and CMMLU, we select STEM-related subsets to specifically evaluate the model's reasoning capabilities in STEM fields, as detailed in \cref{a:benchmarks}.

\subsubsection{Evaluation Setting}
\label{sec:3.3.2}

To ensure statistical robustness and mitigate sampling variability, we conduct $N$ independent generations for each test instance using zero-shot evaluation, adhering to the generation configurations detailed in \cref{tab: gen config} by default. For reasoning-intensive benchmarks with a limited number of test instances, such as AIME2024, AIME2025, BeyondAIME, HMMT\_Feb\_2025, and GPQA-Diamond, we perform $16$ generations per case. In contrast, for elementary benchmarks or those with a larger number of test instances, we use up to $4$ generations per instance. Consequently, we report \textit{Pass@1} as the primary evaluation metric, with additional \textit{Pass@K} (also referred to as \textit{Best@N}) and majority-vote accuracy (\textit{Majority@N}) as supplementary metrics. Further details of evaluation settings are described in \cref{a:evaluation}.

\subsubsection{General Evaluation Results}
\label{sec:3.3.3}

As shown in \cref{tab:3,tab:4}, Logics-STEM-8B-RL significantly outperforms other reasoning models of comparable size from the community, scoring 90.42\% on AIME2024, 87.08\% on AIME2025, 74.79\% on HMMT2025, 62.5\% on BeyondAIME and 73.93\% on GPQA-Diamond. The consistently strong performance across diverse benchmarks further evidences the robust domain-generalization capacity of the proposed data curation strategy, demonstrating the effectiveness of our SFT and RLVR strategies. 

Notably, Logics-STEM-8B-SFT achieves reasoning performance comparable to state-of-the-art open-source reinforcement learning models of similar scale, such as Klear-Reasoner-8B. This finding demonstrates our ability to fully leverage high-quality reasoning datasets and further suggests that, for models of small size, knowledge distillation through SFT can be as effective as RL-centric post-training.

\begin{table}[htbp]
\centering
\footnotesize
\setlength{\tabcolsep}{2pt} 
\caption{The average \textit{pass@1} performance reported on mathematical benchmarks. $^{\dagger}$ denotes results from model developers. $\clubsuit$ denotes a maximum inference budget of 64k context. The best score on a given benchmark is marked in \textbf{bold}, and the second best is $\underline{underlined}$.}
\label{tab:3}
\begin{tabularx}{\textwidth}{X*{6}{c}}
\toprule
\multirow{2}{*}{\textbf{Model}} & \multicolumn{6}{c}{\textbf{Benchmark}} \\
\cmidrule(lr){2-7}
& \textbf{AIME2024} & \textbf{AIME2025} & \textbf{BeyondAIME} & \textbf{HMMT2025} & \textbf{BRUMO2025} & \textbf{MATH500} \\
& avg@16 & avg@16 & avg@16 & avg@16 & avg@16 & avg@4 \\
\midrule 
Qwen3-8B & 76.0$^{\dagger}$ & 67.3$^{\dagger}$ & 42.88 & 44.17 & 67.92 & 97.4$^{\dagger}$ \\
\quad \textit{w/ 64K Context} $\clubsuit$ & 75.62 & 69.17 & 43.88 & 42.29 & 68.12 & 96.3 \\
OpenThinker3-7B & 69.0$^{\dagger}$ & 53.3$^{\dagger}$ & 34.69 & 42.7$^{\dagger}$ & 61.88 & 90.0$^{\dagger}$ \\
AceReason-1.1-7B & 72.6$^{\dagger}$ & 64.8$^{\dagger}$ & 44.69 & 41.88 & 69.79 & 96.35 \\
R1-0528-Qwen3-8B & 77.18 & 68.75 & 42.50 & 49.17 & 70.62 & 94.35 \\
\quad \textit{w/ 64K Context} $\clubsuit$ & 86.0$^{\dagger}$ & 76.3$^{\dagger}$ & 48.69 & 61.5$^{\dagger}$ & 74.17 & 95.15 \\
Klear-Reasoner-8B-SFT & 75.6$^{\dagger}$ & 70.1$^{\dagger}$ & 32.06 & 57.6$^{\dagger}$ & 68.96 & 94.8 \\
\quad \textit{w/ 64K Context} $\clubsuit$ & 82.29 & 81.88 & 52.31 & 67.08 & 81.46 & 95.25 \\
Klear-Reasoner-8B & 83.2$^{\dagger}$ & 75.6$^{\dagger}$ & 48.81 & 60.3$^{\dagger}$ & 72.71 & 95.7 \\
\quad \textit{w/ 64K Context} $\clubsuit$ & \textbf{90.5}$^{\dagger}$ & 83.2$^{\dagger}$ & 50.25 & 70.8$^{\dagger}$ & 77.08 & 95.3 \\
\midrule
Logics-STEM-8B-SFT & 80.62 & 73.33 & 44.31 & 57.71 & 72.92 & 97.5 \\
\quad \textit{w/ 64K Context} $\clubsuit$ & 90.42 & \underline{85.62} & \underline{61.25} & \underline{71.46} & \textbf{86.67} & \textbf{98.85} \\
Logics-STEM-8B-RL & 80.62 & 74.79 & 48.31 & 57.29 & 75.83 & 97.8 \\
\quad \textit{w/ 64K Context} $\clubsuit$ & \underline{90.42} & \textbf{87.08} & \textbf{62.5} & \textbf{74.79} & \underline{86.46} & \underline{98.4} \\
\bottomrule 
\end{tabularx}
\end{table}

\begin{table}[htbp]
\centering
\footnotesize
\setlength{\tabcolsep}{4pt} 
\caption{The average \textit{pass@1} performance reported on STEM benchmarks. $^{\dagger}$ denotes results from model developers. $\clubsuit$ denotes a maximum inference budget of 64k context. The best score on a given benchmark is marked in \textbf{bold}, and the second best is $\underline{underlined}$.}
\label{tab:4}
\begin{tabularx}{\textwidth}{X*{4}{c}}
\toprule
\multirow{2}{*}{\textbf{Model}} & \multicolumn{4}{c}{\textbf{Benchmark}} \\
\cmidrule(lr){2-5}
& \textbf{GPQA-Diamond} & \textbf{R-Bench} & \textbf{MMLU-Pro-STEM} & \textbf{CMMLU-STEM} \\
& avg@16 & avg@1 & avg@1 & avg@1 \\
\midrule 
Qwen3-8B & 62.00$^{\dagger}$ & 64.03 & 83.02 & 85.69  \\
\quad \textit{w/ 64K Context} $\clubsuit$ & 60.23 & 63.35 & 82.99 & 85.92 \\
OpenThinker3-7B & 53.70$^{\dagger}$ & 54.71 & 70.15 & 72.96 \\
AceReason-1.1-7B & 50.09 & 57.08 & 72.52 & 72.21\\
R1-0528-Qwen3-8B & 58.30 & 62.48 & 79.41 & 85.02 \\
\quad \textit{w/ 64K Context} $\clubsuit$ & 61.10$^{\dagger}$ & 63.21 & 80.57 & 85.02 \\
Klear-Reasoner-8B-SFT & 62.12 & 57.04 & 79.90 & 81.80\\
\quad \textit{w/ 64K Context} $\clubsuit$ & 63.64 & 64.03 & 81.62 & 82.62 \\
Klear-Reasoner-8B & 44.76 & 46.36 & 59.80 & 63.90 \\
\quad \textit{w/ 64K Context} $\clubsuit$ & 43.62 & 49.31 & 60.16 & 64.12 \\
\midrule
Logics-STEM-8B-SFT & 72.70 & 72.21 & 85.20 & 85.06 \\
\quad \textit{w/ 64K Context} $\clubsuit$ & \underline{73.11} & \underline{72.12} & \underline{85.24} & \underline{86.82} \\
Logics-STEM-8B-RL & \textbf{74.31} & 71.57 & 85.33 & 87.87 \\
\quad \textit{w/ 64K Context} $\clubsuit$ &73.93 & \textbf{73.67} & \textbf{85.63} & \textbf{88.31} \\
\bottomrule 
\end{tabularx}
\end{table}

Moreover, as evidenced by the substantial improvements in the \textit{Majority@N} on competition-level benchmarks shown in \cref{tab:5}, the application of RLVR after supervised fine-tuning (SFT) significantly enhances model reasoning ability by further concentrating the probability distribution around correct answers, even when SFT has already established a strong baseline.

\begin{table}[htbp]
\centering
\footnotesize
\setlength{\tabcolsep}{2pt} 
\caption{Supplementary performance measured by \textit{Majority@N} and \textit{Best@N} on key reasoning benchmarks. $\clubsuit$ denotes a maximum inference budget of 64k context. The best score on a given benchmark is marked in \textbf{bold}, and the second best is $\underline{underlined}$.}
\label{tab:5}
\begin{tabularx}{\textwidth}{X*{8}{c}}
\toprule
\multirow{3}{*}{\textbf{Model}} & \multicolumn{8}{c}{\textbf{Benchmark}} \\
\cmidrule(lr){2-9}
& \multicolumn{2}{c}{\textbf{AIME2025}} & \multicolumn{2}{c}{\textbf{HMMT2025}} & \multicolumn{2}{c}{\textbf{BeyondAIME}} & \multicolumn{2}{c}{\textbf{GPQA-Diamond}}\\
& Maj@16 & Best@16 & Maj@16 & Best@16 & Maj@16 & Best@16 & Maj@16 & Best@16 \\
\midrule 
Qwen3-8B & 76.67 & 83.33 & 53.33 & 66.67 & 46.00 & 65.00 & 64.14 & 83.33 \\
\quad \textit{w/ 64K Context} $\clubsuit$ & 83.33 & 83.33 & 56.67 & 66.67 & 50.00 & 66.00 & 62.12 & 84.85 \\

R1-0528-Qwen3-8B & 73.33 & 86.67 & 53.33 & 76.67 & 48.00 & 68.00 & 64.65 & 89.39 \\
\quad \textit{w/ 64K Context} $\clubsuit$ & 76.67 & 86.67 & 63.33 & 73.33 & 58.00 & 72.00 & 65.66 & 90.91 \\

Klear-Reasoner-8B-SFT & 73.33 & 83.33 & 63.33 & 80.00 & 34.00 & 59.00 & 66.67 & 90.91 \\
\quad \textit{w/ 64K Context} $\clubsuit$ & 86.67 & \textbf{96.67} & \underline{76.67} & \underline{86.67} & 61.00 & 78.00 & 68.69 & 91.41 \\

Klear-Reasoner-8B & 80.00 & \underline{93.33} & 73.33 & \underline{86.67} & 57.00 & 75.00 & 46.46 & 77.78 \\
\quad \textit{w/ 64K Context} $\clubsuit$ & 83.33 & \textbf{96.67} & 73.33 & 83.33 & 59.00 & 74.00 & 46.46 & 79.80 \\
\midrule

Logics-STEM-8B-SFT & 80.00 & 90.00 & 60.00 & 76.67 & 46.00 & 68.00 & \underline{76.26} & 91.41 \\
\quad \textit{w/ 64K Context} $\clubsuit$ & \underline{90.00} & \underline{93.33} & \underline{76.67} & \textbf{93.33} & \underline{66.00} & \underline{80.00} & 75.25 & 91.41 \\

Logics-STEM-8B-RL & 80.00 & 90.00 & 63.33 & 73.33 & 50.00 & 67.00 & \underline{76.26} & \underline{91.92} \\
\quad \textit{w/ 64K Context} $\clubsuit$ & \textbf{93.33} & \textbf{96.67} & \textbf{83.33} & \textbf{93.33} & \textbf{67.00} & \textbf{82.00} & \textbf{77.27} & \textbf{92.42} \\
\bottomrule 
\end{tabularx}
\end{table}


\subsection{Ablation Study of Data}
\label{sec: data ablation}

\subsubsection{Sampling Strategy}
\label{sec: sampling strsategy}

As explained in \cref{sec:stratified sampling}, further sampling over the massive outputs of the curation pipeline is essential. Therefore we conduct a series of ablation experiments to identify the most effective sampling strategy. Specifically, for each sampling strategy, we uniformly sample 200K records to construct the training set and fine-tune the model for one epoch.

\paragraph{Stratified Sampling.}
Length-based sampling has been demonstrated to be an effective sampling method \citep{guha2025openthoughts}; nevertheless, our experiments demonstrate that its exclusive use degrades the model’s elementary reasoning and generalization capacities, as reported in \cref{tab: ablate}. the ablation model trained solely with length-based sampling incurs significant performance drops on MATH500, MMLU-Pro-Math, and GPQA-Diamond, despite notable gains on AIME2024, AIME2025, and HMMT2025.

\paragraph{Difficulty-Based Sampling.}
Another set of ablation experiments focuses on variants of stratified sampling. We explore sampling based on the token length of response generated by the teacher model and the annotated education level of the question. The results in Table \ref{tab: ablate} demonstrate that the former serves as a natural and more effective proxy for question difficulty.

We therefore adopt length-based stratified sampling to mitigate catastrophic forgetting of elementary reasoning capabilities, while scaling the training set to strengthen advanced reasoning capabilities.

\begin{table}[htbp]
\centering
\footnotesize
\setlength{\tabcolsep}{2pt} 
\caption{\centering Ablation performance based on Qwen3-8B across key reasoning benchmarks}
\label{tab: ablate}
\begin{tabularx}{\textwidth}{X*{5}{c}}
\toprule
\multirow{2}{*}{\textbf{Model}} & \multicolumn{5}{c}{\textbf{Benchmark}} \\
\cmidrule(lr){2-6}
& \textbf{AIME2025} & \textbf{HMMT2025} & \textbf{MATH500} & \textbf{GPQA-D} & \textbf{MMLU-Pro-STEM} \\
& avg@16 & avg@16 & avg@4 & avg@16 & avg@1 \\
\midrule 
Purely Length-Based Sampling & \textbf{55.62} & \textbf{36.88} & $\underline{95.35}$ & 46.88 & 75.81 \\
Stratified Sampling \\
\quad \textit{w/ Length-Based} & $\underline{53.12}$ & 34.38 & \textbf{95.5} & \textbf{49.91} & \textbf{79.15} \\
\quad \textit{w/ Annotation-Based} & 52.29 & $\underline{34.58}$ & 94.2 & $\underline{48.42}$ & $\underline{77.82}$ \\
\bottomrule 
\end{tabularx}
\end{table}

\subsubsection{Source of failure-driven Synthetic Data}
To investigate how different types of benchmark-recall document synthesis affect second-stage training, we construct training data using two representative benchmarks:
(i) Scientific QA (e.g., GPQA),
(ii) Mathematical reasoning (e.g., Math500, AIME2024/2025, and HMMT).
Following a unified data construction pipeline, we generate second-stage training samples from incorrectly answered questions in these benchmarks and train models independently. Experimental results demonstrate that training data synthesized from GPQA errors not only significantly improves GPQA evaluation performance but also delivers notable generalization gains across mathematical reasoning benchmarks (e.g., AIME, HMMT). In contrast, training data derived solely from mathematical benchmarks yields limited improvement in overall STEM capabilities. 

\begin{table}[htbp]
\centering
\footnotesize
\setlength{\tabcolsep}{2pt} 
\caption{\centering Impact of Synthetic Data Sources}
\label{tab: data_source_ablate}
\begin{tabularx}{\textwidth}{X*{6}{c}}
\toprule
\multirow{2}{*}{\textbf{Algorithm}} & \multicolumn{6}{c}{\textbf{Benchmark}} \\
\cmidrule(lr){2-7}
& \textbf{AIME2024} & \textbf{AIME2025} & \textbf{HMMT2025} & \textbf{GPQA-D} & \textbf{MMLU-Pro-STEM} & \textbf{CMMLU-STEM} \\
& Avg@16 & Avg@16 & Avg@16 & Avg@16 & Avg@1 & Avg@1 \\
\midrule 
baseline & 90.21 & 85.56 & 68.96 & 73.20 & 85.35 & 86.82 \\
GRPO\\
\quad sci-syn-data & \textbf{90.42} & \textbf{87.08} & \textbf{74.79} & \textbf{73.93} & \textbf{85.63} & \textbf{88.31} \\
\quad math-syn-data & 89.79 & 86.04 & 70.42 & 72.70 & 85.10 & 87.27 \\
\bottomrule 
\end{tabularx}
\end{table}

This finding suggests that the complex reasoning patterns inherent in scientific QA tasks may possess stronger cross-domain transferability, making them more valuable for enhancing general reasoning abilities.

\subsection{Ablation Study of Algorithms}
\label{sec: algorithm ablation}
In this section, we examine how different settings and modules influence the RLVR training process. We present ablation experiments with detailed results and demonstrate how they inform the design of our final training recipe.

\subsubsection{RLVR Algorithm}

Our experiments cover different loss functions, including Group Relative Policy Optimization (GRPO) (\citet{shao2024deepseekmathpushinglimitsmathematical}), Decoupled Clip and Dynamic Sampling Policy Optimization (DAPO) (\citet{yu2025dapoopensourcellmreinforcement}).



We first conduct ablation studies on Qwen2.5-7B-Instruct using the DAPO-Math-17K dataset and find that employing the DAPO objective in conjunction with Reinforce++ as the advantage estimator yields more stable and consistent improvements.
As shown in \cref{tab: algo_ablate}, an early ablation indicates that the DAPO loss function outperforms the GRPO Loss on our SFT model. 
Therefore, the DAPO objective coupled with Reinforce++ as the advantage estimator is adopted as the final algorithm.

\begin{table}[htbp]
\centering
\footnotesize
\setlength{\tabcolsep}{2pt} 
\caption{Early Ablation Results based on Logics-STEM-8B-SFT across key reasoning benchmarks}
\label{tab: algo_ablate}
\begin{tabularx}{\textwidth}{X*{6}{c}}
\toprule
\multirow{2}{*}{\textbf{Algorithm}} & \multicolumn{6}{c}{\textbf{Benchmark}} \\
\cmidrule(lr){2-7}
& \textbf{AIME2024} & \textbf{AIME2025} & \textbf{HMMT2025} & \textbf{GPQA-D} & \textbf{MMLU-Pro-STEM} & \textbf{CMMLU-STEM} \\
& Avg@16 & Avg@16 & Avg@16 & Avg@16 & Avg@1 & Avg@1 \\
\midrule 
DAPO & 90.83 & \textbf{85.62} & \textbf{72.92} & \textbf{73.96} & 85.23 & \textbf{87.34} \\
GRPO & \textbf{91.46} & 83.54 & 69.58 & 73.55 & \textbf{85.29} & \textbf{87.34} \\
\bottomrule 
\end{tabularx}
\end{table}

\subsubsection{Dynamic Sampling}

Following \citet{yu2025dapoopensourcellmreinforcement}, we take advantage of dynamic data sampling strategy in our training pipeline. Specifically, we first drop the responses that are longer than our default maximum length, it is called overlong length filter. Also, for the difficulty filter, for each prompt, we will calculate the average score within the group of all its responses, if it is lower or greater than our set threshold, the data will not be involved in the calculation of loss function. 

In this experiment, we conduct two groups of trials: one using our original setting (with both the overlong-length and difficulty filters enabled), and the other with both filters disabled. We set the lower and upper thresholds for the average response score to 0.1 and 0.95, respectively.

\begin{figure}[htbp]
  \centering
  \includegraphics[width=0.48\linewidth]{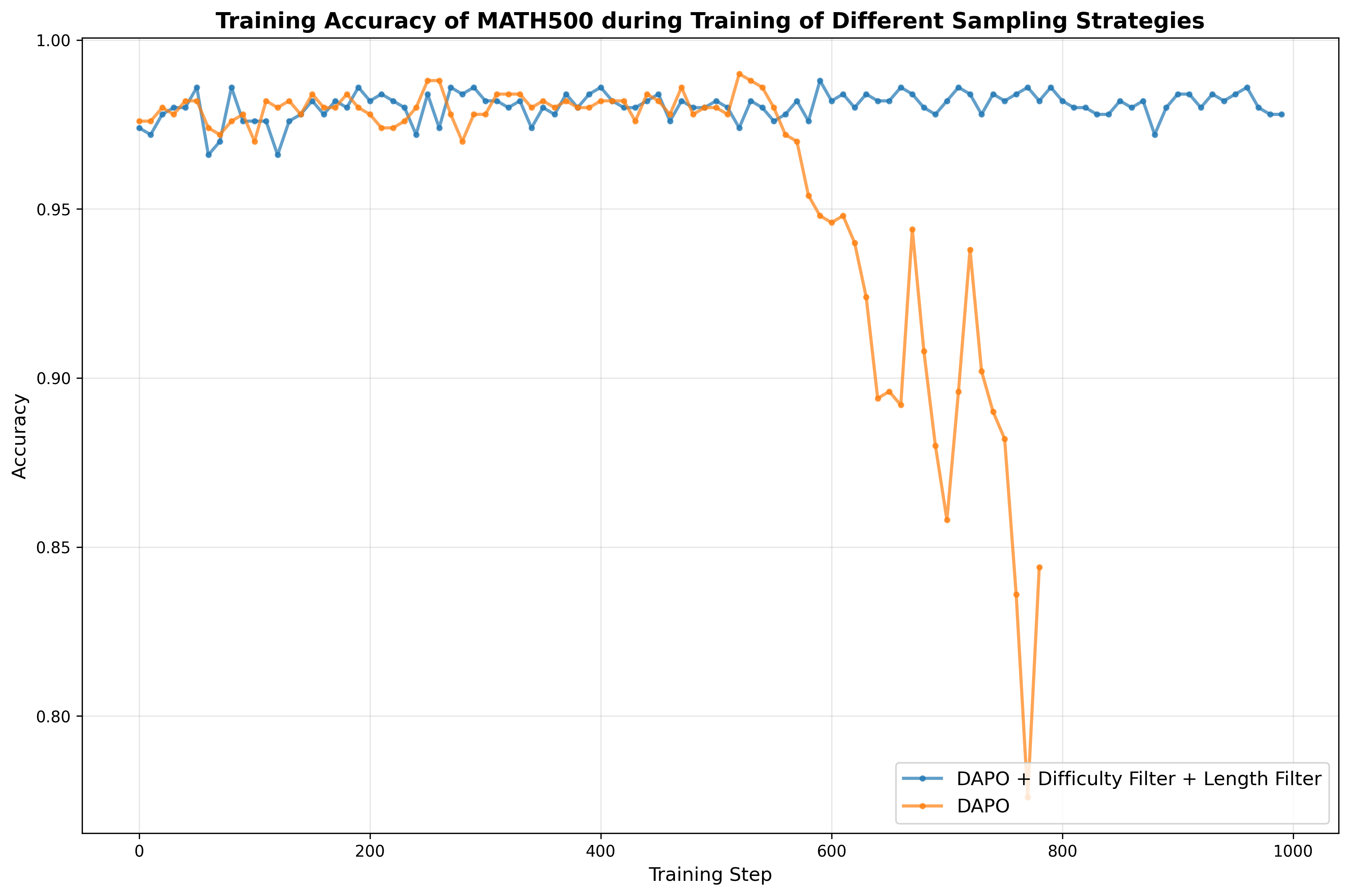}
  \hfill 
  \includegraphics[width=0.48\linewidth]{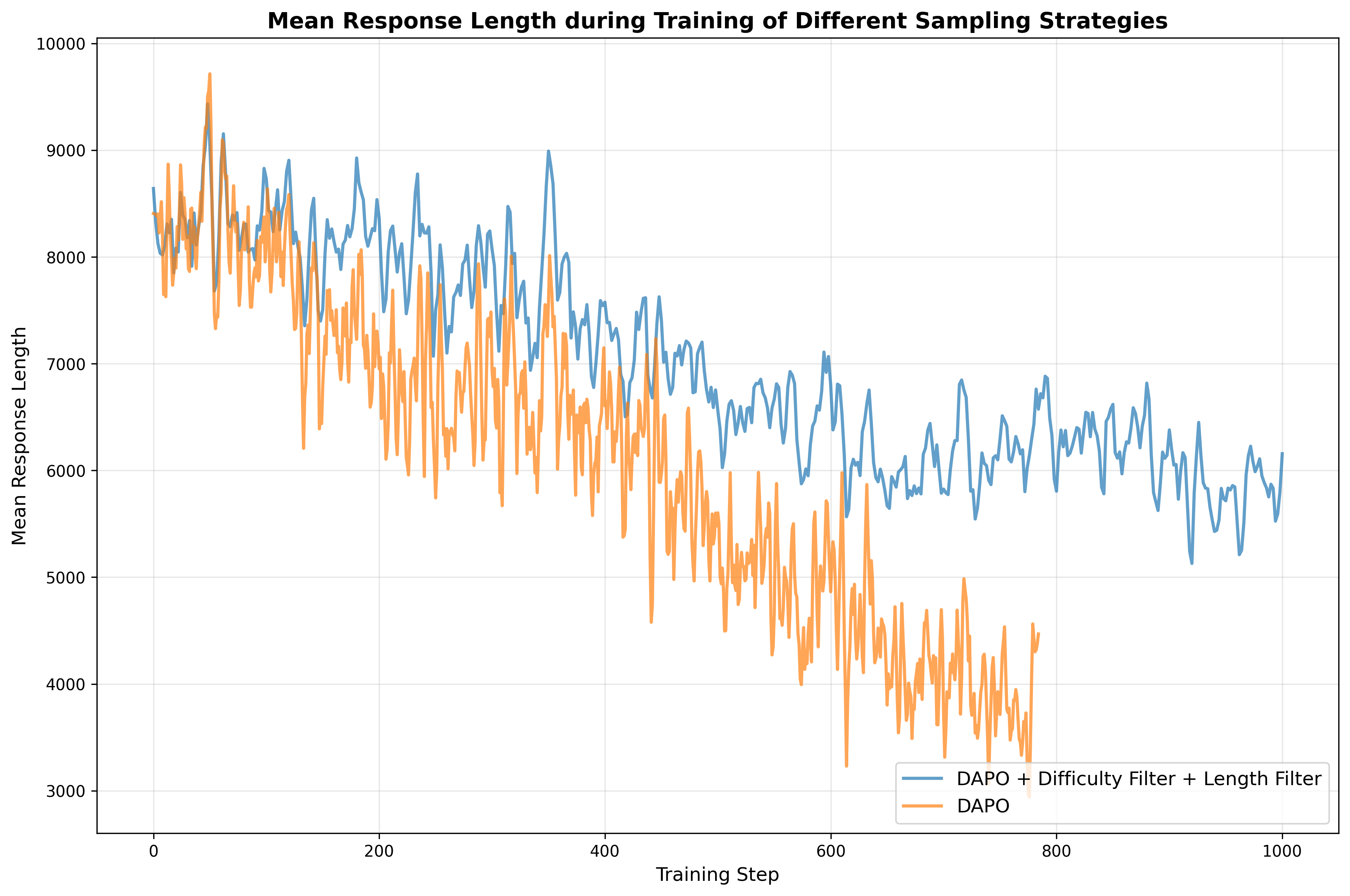}
  
  \caption{Left: Training Accuracy on MATH500 with Different Filter Strategies; Right: Mean Response Length with Different Filter Strategies.}
  \label{fig: filter_contrast}
\end{figure}

As we can observe in \cref{fig: filter_contrast}, the group without a filter tends to collapse in both the training accuracy and response length in an early stage. 

Our analysis indicates that response groups with average scores that are either too high or too low yield near-zero advantage values after normalization, which in turn results in negligible or zero gradient updates. Intuitively, such samples are either too “easy” or too “hard” for the current model and provide little useful signal for learning. Additionally, without the overlong-length filter, responses exceeding the maximum allowed length are truncated, making it difficult to extract the final output and leading to a short response length. Consequently, both the overlong-length filter and the difficulty filter are essential to ensure stable and effective training.


\subsubsection{Length-based Reward and Repetition Penalty}

\begin{table}[htbp]
\centering
\footnotesize
\setlength{\tabcolsep}{1pt} 
\caption{\centering Impact of Length Reward and Repetition Penalty on Results}
\label{tab: len_reward_ablate}
\begin{tabularx}{\textwidth}{X*{6}{c}}
\toprule
\multirow{2}{*}{\textbf{Algorithm}} & \multicolumn{6}{c}{\textbf{Benchmark}} \\
\cmidrule(lr){2-7}
& \textbf{AIME2024} & \textbf{AIME2025} & \textbf{HMMT2025} & \textbf{GPQA-D} & \textbf{MMLU-Pro-STEM} & \textbf{CMMLU-STEM} \\
& Avg@16 & Avg@16 & Avg@16 & Avg@16 & Avg@1 & Avg@1 \\
\midrule 
GRPO \\
\quad w/ len reward & \textbf{90.42} & \textbf{87.08} & \textbf{74.79} & \textbf{73.93} & \textbf{85.63} & \textbf{88.31} \\
\quad w/o len reward & 90.21 & 85.00 & 70.83 & 73.74 & 84.87 & 85.84 \\
\bottomrule 
\end{tabularx}
\end{table}

Experimental results demonstrate that incorporating the reasoning-length reward and repetition penalty mechanisms yields significant performance improvements ranging from 2 to 3 percentage points on benchmark datasets, including AIME2025, HMMT, and CMMLU. Consistent, albeit modest, gains are also observed across all other evaluated benchmarks.

\subsection{The Generalizability of Data Engine}

\subsubsection{Generalizability of Logics-STEM-SFT-Dataset to Larger-scaled Models}
In addition to the superior performance achieved by the model fine-tuned from Qwen3-8B (as presented in \cref{sec:3.3.3}), we also experiment with several other models as initial models during the supervised fine-tuning (SFT) stages. All of these experiments further confirm the effectiveness of our data engine, as shown in~\cref{tab: generalization of long cot}.

Firstly, we fine-tune Qwen3-8B-Base with the identical Logics-STEM-SFT-2.2M dataset. The resulting model performs on par with, though marginally weaker than, Logics-STEM-8B-SFT while substantially surpassing Qwen3-8B. These gains demonstrate that our dataset is sufficiently diverse and of high quality to support full-scale supervised fine-tuning (SFT) in the STEM domain.

Additionally, we evaluate the scalability of our approach by fine-tuning Qwen3-32B. Remarkably, the resulting model, Logics-STEM-32B-SFT, still exhibits substantial improvements in reasoning performance, further corroborating the high quality of our dataset.

\begin{table}[htbp]
\centering
\footnotesize
\setlength{\tabcolsep}{2pt} 
\caption{Generalizability of Long CoT data measured by average \textit{pass@1} performance reported on key reasoning benchmarks. $\clubsuit$ denotes a maximum inference budget of 64k context.}
\label{tab: generalization of long cot}
\begin{tabularx}{\textwidth}{X*{6}{c}}
\toprule
\multirow{2}{*}{\textbf{Model}} & \multicolumn{6}{c}{\textbf{Benchmark}} \\
\cmidrule(lr){2-7}
& \textbf{AIME2024} & \textbf{AIME2025} & \textbf{BeyondAIME} & \textbf{HMMT2025} & \textbf{MATH500} & \textbf{GPQA-D} \\
& avg@16 & avg@16 & avg@16 & avg@16 & avg@4 & avg@16 \\

\midrule %
Qwen3-8B-Base-SFT & 81.04 & 72.92 & 43.62 & 57.5 & 96.85 & 72.79 \\
\quad \textit{w/ 64K Context} & 88.75 & 83.33 & \textbf{61.38} & 68.96 & 98.45 & \textbf{73.17} \\
Logics-STEM-8B-SFT & 80.62 & 73.33 & 44.31 & 57.71 & 97.5 & 72.70 \\
\quad \textit{w/ 64K Context} $\clubsuit$ & \textbf{90.42} & \textbf{85.62} & 61.25 & \textbf{71.46} & \textbf{98.85} & 73.11 \\

\midrule 

Qwen3-32B$\clubsuit$ & 82.08 & 73.12 & 50.00 & 55.00 & 96.55 & 68.18 \\
Logics-STEM-32B-SFT$\clubsuit$ & \textbf{92.71} & \textbf{89.38} & \textbf{63.56} & \textbf{77.92} & \textbf{98.30} & \textbf{76.64} \\

\bottomrule 
\end{tabularx}
\end{table}

\subsubsection{Generalizability of Failure-Driven Data}
In addition to the RLVR experiments, we conduct continual SFT experiments using failure-driven synthetic data to assess the universal effectiveness of the data across different post-training paradigms. Analogous to the RLVR training setup, we conduct experiments on Logics-8B-SFT, with training hyperparameters detailed in the Appendix.
To mitigate catastrophic forgetting, we experimentally construct mixtures of 34K failure-driven synthetic data and existing SFT training data at varying ratios. Notably, under an approximately 1:1 mixture of failure-driven synthetic data and existing SFT data, the model achieves improvements comparable to those obtained with the RLVR approach, as presented in \cref{tab: generalization of failure-driven} .

\begin{table}[htbp]
\centering
\footnotesize
\setlength{\tabcolsep}{2pt} 
\caption{Generalizability of Failure-Driven data measured by average \textit{pass@1} performance reported on key reasoning benchmarks. $\clubsuit$ denotes a maximum inference budget of 64k context. Exp1 represents continual SFT with 34K pure failure-driven synthetic data. Exp2 represents continual SFT with mixtures of 34K failure-driven synthetic data and 10K original SFT data. Exp3 represents continual SFT with mixtures of 34K failure-driven synthetic data and 30K original SFT data.} 
\label{tab: generalization of failure-driven}
\begin{tabularx}{\textwidth}{X*{6}{c}}
\toprule
\multirow{2}{*}{\textbf{Model}} & \multicolumn{6}{c}{\textbf{Benchmark}} \\
\cmidrule(lr){2-7}
& \textbf{AIME2024} & \textbf{AIME2025} & \textbf{BeyondAIME} & \textbf{HMMT2025} & \textbf{MATH500} & \textbf{GPQA-D} \\
& avg@16 & avg@16 & avg@16 & avg@16 & avg@4 & avg@16 \\

\midrule 
Logics-STEM-8B-SFT  $\clubsuit$ & 90.42 & \underline{85.62} & 61.25 & 71.46 & \textbf{98.85} & 73.11 \\
Logics-STEM-8B-RL  $\clubsuit$ & \underline{90.42} & \textbf{87.08} & \underline{62.50} & \textbf{74.79} & \underline{98.40} & \textbf{73.93} \\
\midrule 
Logics-STEM-8B-Exp1 $\clubsuit$ & 88.75 & 83.12 & 59.69 & 71.04 & 97.00 & 70.96\\
Logics-STEM-8B-Exp2 $\clubsuit$ & 90.83 & 84.79 & 62.06 & 72.08 & 97.35 & 72.19\\
Logics-STEM-8B-Exp3 $\clubsuit$ & \textbf{91.88} & \underline{85.62} & \textbf{62.69} & \underline{73.33} & 97.40 & \underline{73.33}\\

\bottomrule 
\end{tabularx}
\end{table}

\subsection{Scale Inference Budget}
\label{sec: Scale Inference Budget}
Due to different evaluation settings in officially reported results that limit direct comparisons across models, we systematically assess the effect of inference budget by independently evaluating each model under varying context settings. 

As illustrated in \cref{tab:3,tab:4}, scaling up inference budget from 32K context to 64K context leads to significant improvements on competition-level benchmarks such as AIME2024, AIME2025, and HMMT2025, whereas improvements on elementary benchmarks remain modest. The observation is consistent with the intuition that more challenging problems require deeper deliberation. Moreover, reasoning-oriented models typically benefit more from the additional inference budget than the vanilla Qwen3-8B baseline.

\subsection{Unsuccessful Attempts}
\label{sec: unsuccessful attempts}
\paragraph{Numeric-Answer-Only RLVR} In an experiment, we apply reinforcement learning on \textit{DAPO-Math-17k}, a dataset that exclusively contains numeric verifiable answers. We observe a marked deterioration in the model’s instruction-following ability, specifically in its capacity to respond with option letters as instructed, which leads to significant performance degradation on GPQA-Diamond, R-Bench, and MMLU-Pro-STEM. This observation explains the poor performance of Klear-Reasoner-8B on these benchmarks (See \cref{tab:4}).


\paragraph{Entropy Loss in RLVR} In our ablation study, we compare training objectives with and without an entropy loss term, and we also explore adaptive entropy control following the approach of \citet{he2025skyworkopenreasoner1}, in which we stick to their hyperparameter setting. Our experiments show that either incorporating a naive entropy loss or implementing adaptive entropy loss requires careful hyperparameter tuning and is prone to causing entropy explosion, as illustrated in \cref{fig: entr_contrast}. Consequently, we exclude this term from our final training recipe. 

\paragraph{Scale Failure-Driven Synthetic Data} We further investigate scaling the volume of failure-driven synthetic data by relaxing the relevance threshold to retrieve more educational documents associated with each failure case. However, as detailed in \cref{{appendix: scale failure-driven}}, this actually degrades the model’s performance. We hypothesize that the additional documents are less strongly related to the original failures, thereby breaking the expected scaling behavior.

\section{Conclusion}
\label{sec:4}
This technical report introduces Logics-STEM, a state-of-the-art reasoning model specifically designed for STEM-related domains. Through proposed data curation pipeline combined with the difficulty-based stratified sampling strategy, we effectively select high-quality reasoning data from both open-access sources and synthetic datasets, thereby ensuring robust supervised fine-tuning. The further integration of failure-driven post-training and complementary dynamic sampling leads to substantial improvements in reasoning performance. For future work, we plan to further investigate reinforcement learning techniques, scale the model to larger architectures, and extend its capabilities to a broader range of tasks, including coding and tool-integrated reasoning.


\bibliography{ref}

\appendix
\section{More about Data}

\subsection{Source of Data}
\label{a:source}
We collect questions from the following highly regarded and frequently cited datasets: NuminaMath1.5 (\citet{numina_math_datasets}), OpenThoughts3 (\citet{guha2025openthoughts}), Mixture-of-Thoughts (\citet{openr1}), AceReason-1.1-SFT (\citet{liu2025acereason}), AceReason-Math (\citet{chen2025acereason}), OpenScienceReasoning-2 \footnote{https://huggingface.co/datasets/nvidia/OpenScienceReasoning-2}, OpenMathReasoning (\citet{moshkov2025aimo}), Llama-Nemotron-Post-Training-Dataset (\citet{bercovich2025llama}), DLR-Web (\citet{liu2025designer}), DLR-Book (\citet{liu2025designer}), Skywork-OR1-RL-Data (\cite{he2025skywork}), NaturalReasoning (\citet{yuan2025naturalreasoning}), DeepMath-103K (\citet{he2025deepmath}), DAPO-Math-17k (\citet{yu2025dapo}), TheoremQA (\citet{chen2023theoremqa}), JEEBench (\citet{arora-etal-2023-llms}), GPQA-Main (\citet{rein2024gpqa}), GSM8K (\citet{cobbe2021training}), AIME \footnote{https://huggingface.co/datasets/di-zhang-fdu/AIME\_1983\_2024}, AMC \footnote{https://huggingface.co/datasets/kaggle-aimo/amc\_filtered}, s1-teasers (\citet{muennighoff2025s1}), s1-probs (\citet{muennighoff2025s1}), openaimath (\citet{muennighoff2025s1}).

\subsection{Annotation}
\label{a:annotation}

\begin{flushleft}
We structure and organize our reasoning data according to the following dimensions:
\begin{itemize}
    \item Questions are considered invalid and excluded from subsequent curation if they: (1) lack required images; (2) is incomplete; (3) is unsolvable. 
    \item Data are categoried into 6 domains: Math, STEM(excluding Math), Humanities\&SocialScience, Business, Medicine\&Biology, Code.
    \item Data are categorized by educational level into six tiers: elementary, junior-secondary, senior-secondary, undergraduate, graduate, and competition.
    \item Answer types are taxonomized as Boolean, multiple-choice, numeric, vector/matrix, interval, expression, string, proof, textual explanation, or other. In a narrow sense, verifiable answers include only multiple-choice and numeric types, while in a broader sense, verifiable answers further encompass Boolean, vector/matrix, interval, and expression types.
\end{itemize}
\end{flushleft}

\subsection{Generation Configuration}
\label{a:generation}

We adopt the recommended configuration parameters provided by Qwen team for both response distillation and evaluation inference, as detailed in \cref{tab: gen config}.

\begin{table}[htbp]
    \centering
    \caption{\centering Generation configuration for response generation}
    \begin{tabular}{lcc}
        \toprule
        Parameter & Value \\
        \midrule
        Temperature  & 0.6 \\
        Top-K & 20 \\
        Top-P & 0.95 \\
        Max Context Length & 32768 \\
        \bottomrule
    \end{tabular}
    \label{tab: gen config}
\end{table}

\subsection{STEM Benchmarks}
\label{a:benchmarks}

\begin{flushleft}
As described in Section 3.3, we select STEM-related subsets from MMLU-Pro and CMMLU to evaluate the model’s reasoning capabilities in STEM domains. \\
\textbf{MMLU-Pro-STEM} is composed of the following categories extracted from MMLU-Pro: "math," "health," "physics," "biology," "chemistry," "computer science," and "engineering." \\
\textbf{CMMLU-STEM} is composed of the following subsets from MMLU-Pro: "high\_school\_mathematics", "elementary\_mathematics", "college\_mathematics", "high\_school\_physics", "high\_school\_chemistry", "high\_school\_biology", "electrical\_engineering", "conceptual\_physics", and "college\_actuarial\_science".
\end{flushleft}

\section{More about Evaluation}

\subsection{Setting}
\label{a:evaluation}

\subsubsection{Zero-shot Evaluation}

To ensure consistency, we employ a uniform zero-shot evaluation protocol across all benchmarks, where the model is instructed to enclose its final answer within \texttt{$ \backslash $boxed\{\}} in response to each question. \\
For benchmarks featuring multiple-choice questions, such as GPQA-Diamond, MMLU-Pro-STEM, CMMLU-STEM, we prompt the model with the following template:

\begin{examplebox}
\textbf{Example} Two quantum states with energies E1 and E2 have a lifetime of $10^-9$ sec and $10^-8$ sec, respectively. We want to clearly distinguish these two energy levels. Which one of the following options could be their energy difference so that they can be clearly resolved?\\

A. $10^-4$ eV\\
B. $10^-11$ eV\\
C. $10^-8$ eV\\
D. $10^-9$ eV\\

Please put your final option letter within \$$\backslash$boxed\{\}\$.
\end{examplebox}

For other benchmarks, the following template is adopted:

\begin{examplebox}
\textbf{Example} Let $p$ be the least prime number for which there exists a positive integer $n$ such that $n^{4}+1$ is divisible by $p^{2}$. Find the least positive integer $m$ such that $m^{4}+1$ is divisible by $p^{2}$.\\

Please put your final answer within \$$\backslash$boxed\{\}\$.
\end{examplebox}

\subsubsection{Answer Verification}
We extract the answer from the \texttt{\textbackslash boxed\{\}} expression of the model’s response and evaluate its correctness by comparing it with the ground-truth answer using the \texttt{math-verify} library.

\subsubsection{Evaluation Metrics}
For each test instance, we calculate three metrics as following:
\begin{itemize}
    \item \textbf{Pass@1} is defined as the average accuracy across $N$ rollout samples.
    \item \textbf{Best@N} is assigned a value of 1 if any of the $N$ rollout samples matches the gold answer, and 0 otherwise.
\item \textbf{Majority@N} is assigned a value of 1 if the majority answer among the $N$ rollout samples matches the gold answer, and 0 otherwise.
\end{itemize}

Consequently, we report the evaluation result of each benchmark  as the average of the evaluation metrics computed over all test instances.

\subsection{Training Setting}
\subsubsection{Hyperparamaters}
Hyperparameters we use are listed in \cref{tab:Hyperparam for SFT,tab: hyperparam for rlvr,tab:Hyperparam for Continual SFT}.
\begin{table}[htbp]
    \centering
    \caption{\centering Hyperparameters for Supervised Fine-Tuning(SFT)}
    \begin{tabular}{lcc}
        \toprule
        Parameter & Value \\
        \midrule
        Batch Size  & 128 \\
        Learning Rate & 4e-5 \\
        Learning Rate Scheduler & Cosine \\
        Warmup Steps & 1000 \\
        Epoch & 3 \\
        \bottomrule
    \end{tabular}
    \label{tab:Hyperparam for SFT}
\end{table}

\begin{table}[htbp]
    \centering
    \caption{\centering Hyperparameters for Reinforcement Learning with Verified Rewards(RLVR)}
    \begin{tabular}{lcc}
        \toprule
        Parameter & Value \\
        \midrule
        Rollout Batch Size  & 64 \\
        Learning Rate & 1e-6 \\
        Number of Return Sequences in Group & 8 \\
        Response Length & 32768 \\
        Low Threshold for Difficulty Mask & 0.1 \\
        High Threshold for Difficulty Mask & 0.95 \\
        \bottomrule
    \end{tabular}
    \label{tab: hyperparam for rlvr}
\end{table}

\begin{table}[htbp]
    \centering
    \caption{\centering Hyperparameters for Continual Supervised Fine-Tuning(SFT)}
    \begin{tabular}{lcc}
        \toprule
        Parameter & Value \\
        \midrule
        Batch Size  & 64 \\
        Learning Rate & 1e-5 \\
        Learning Rate Scheduler & Cosine \\
        Epoch & 3 \\
        \bottomrule
    \end{tabular}
    \label{tab:Hyperparam for Continual SFT}
\end{table}

\subsection{Supplementaries for Ablation Studies in RLVR}

\subsubsection{Adaptive Entropy Control}

\begin{figure}[h]
  \centering
  \includegraphics[width=0.48\linewidth]{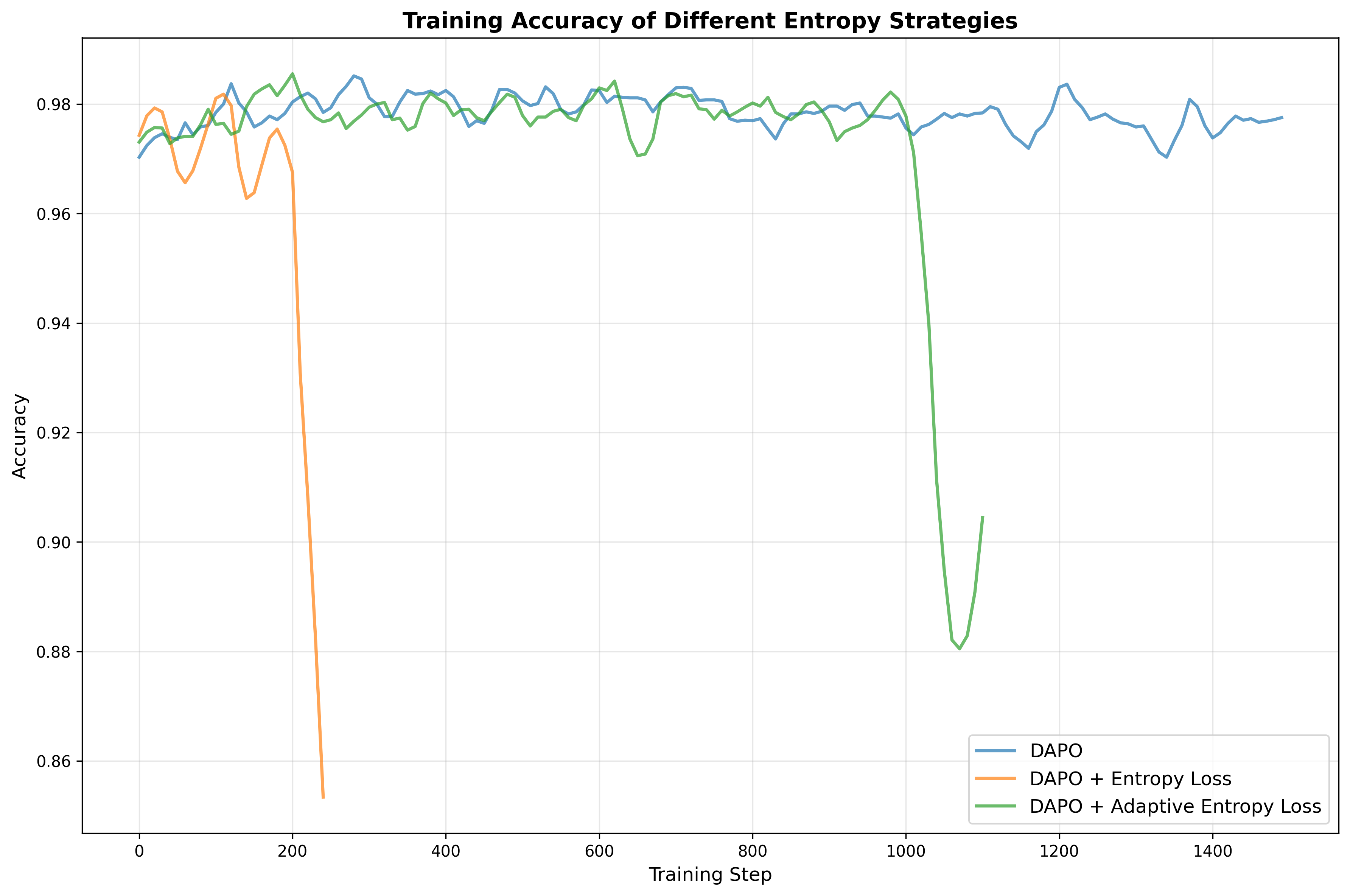}
  \hfill 
  \includegraphics[width=0.48\linewidth]{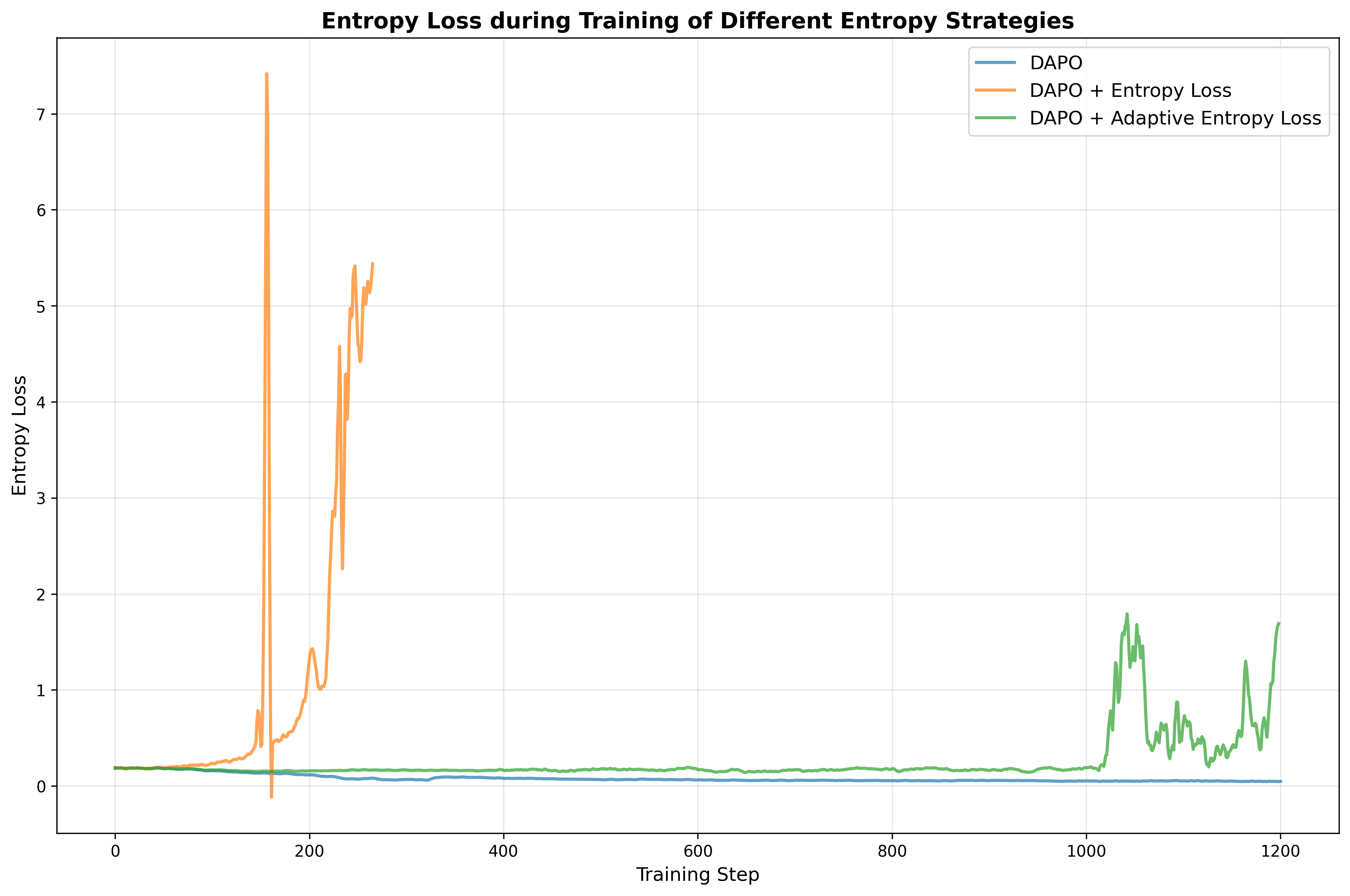}
  
  \caption{Left: Training Accuracy on MATH500 with Different Entropy Strategies; Right: Entropy Loss with Different Entropy Strategies.}
  \label{fig: entr_contrast}
\end{figure}

We experiment with various entropy-based strategies but failed to achieve a stable entropy loss or consistent improvements in accuracy while training, as shown in \cref{fig: entr_contrast}.

\subsubsection{Scale Failure-Driven Synthetic Data}
\label{appendix: scale failure-driven}
As described in \cref{sec: unsuccessful attempts}, the volume of  failure-driven synthetic data is scaled to 150K. Fine-tuning the model on a 1:1 mixture of these 150 K synthetic instances and 150 K existing SFT examples yields the results reported in \cref{tab: scaling of failure-driven}, where we observe gains only on BeyondAIME and GPQA-Diamond accompanied by performance drops on all other benchmarks, despite the increased number of training steps.

\begin{table}[htbp]
\centering
\footnotesize
\setlength{\tabcolsep}{2pt} 
\caption{Generalizability of Failure-Driven data measured by average \textit{pass@1} performance reported on key reasoning benchmarks. $\clubsuit$ denotes a maximum inference budget of 64k context. Exp3 represents continual SFT with mixtures of 34K failure-driven synthetic data and 30K original SFT data. Exp4 represents continual SFT with scaling mixtures of 150K failure-driven synthetic data and 150K existing SFT data.}
\label{tab: scaling of failure-driven}
\begin{tabularx}{\textwidth}{X*{6}{c}}
\toprule
\multirow{2}{*}{\textbf{Model}} & \multicolumn{6}{c}{\textbf{Benchmark}} \\
\cmidrule(lr){2-7}
& \textbf{AIME2024} & \textbf{AIME2025} & \textbf{BeyondAIME} & \textbf{HMMT2025} & \textbf{MATH500} & \textbf{GPQA-D} \\
& avg@16 & avg@16 & avg@16 & avg@16 & avg@4 & avg@16 \\

\midrule 
Logics-STEM-8B-SFT  $\clubsuit$ & 90.42 & \underline{85.62} & 61.25 & 71.46 & \textbf{98.85} & 73.11 \\
Logics-STEM-8B  $\clubsuit$ & \underline{90.42} & \textbf{87.08} & 62.50 & \textbf{74.79} & \underline{98.40} & \underline{73.93} \\
\midrule 
Logics-STEM-8B-Exp3 $\clubsuit$ & \textbf{91.88} & \underline{85.62} & \underline{62.69} & \underline{73.33} & 97.40 & 73.33\\
Logics-STEM-8B-Exp4 $\clubsuit$ & 89.58 & \underline{85.62} & \textbf{64.38} & 71.67 & 97.75 & \textbf{74.84}\\

\bottomrule 
\end{tabularx}
\end{table}

\section{Theoretical Results for Failure-Driven Post-training}
\label{apdx: RL_theory}

\subsection{The Resources of Failure Modes}
\label{apdx: source_failure}

\paragraph{Minizing the loss on $P_0$ may not minimize the target risk on $P^*$.}
Consider a model trained by empirical risk minimization (ERM) on the stage-1 training distribution $P_0$:
\begin{equation}
\label{eq:erm_p0}
\hat{\theta}_0=\arg\min_{\theta}\ \mathbb{E}_{(x,y)\sim P_0}\big[\ell_\theta (x,y)\big].
\end{equation}
However, our real goal is to minimize the risk under the (unknown) target distribution $P^*$:
\begin{equation}
\label{eq:erm_pstar}
\theta^*=\arg\min_{\theta}\ \mathbb{E}_{(x,y)\sim P^*}\big[\ell_\theta(x,y)\big].
\end{equation}
We can decompose $P(x,y)=P(x)P(y\mid x)$, which gives a good interpretation. 

In general, $\hat{\theta}_0 \neq \theta^*$ because the underlying measure in the expectation changes from $P_0$ to $P^*$. This mismatch manifests in several common failure modes:

\textbf{(1) Covariate/domain shift.} The input marginal differs, i.e., $P_0(x)\neq P^*(x)$, so regions that are frequent under $P^*$ may be rarely sampled under $P_0$. When the support overlaps, the target risk admits an importance-weighted form
\begin{equation}
\label{eq:iw}
\mathbb{E}_{(x,y)\sim P^*}\big[\ell_\theta(x,y)\big]
=
\mathbb{E}_{(x,y)\sim P_0}\!\left[\frac{P^*(x,y)}{P_0(x,y)}\,\ell_\theta(x,y)\right],
\end{equation}
which highlights that large density ratios $\frac{P^*(x,y)}{P_0(x,y)}$ correspond to regions that are under-covered by $P_0$ but can dominate the target risk.

\textbf{(2) Support mismatch.} A more severe problem occurs when there exists a region $\mathcal{A}$ such that
\begin{equation}
\label{eq:support_mismatch}
P^*(\mathcal{A})>0,\qquad P_0(\mathcal{A})=0,
\end{equation}
meaning that the target distribution assigns non-zero mass to inputs never seen during stage-1 training. In this case, ERM on $P_0$ cannot provide guarantees on $\mathcal{A}$. This would cause a extremely large density ratio.

\textbf{(3) Underweighting high-loss regions.} Even if $P_0$ and $P^*$ share support, uniform sampling under $P_0$ may assign very small probability mass to difficult, high-loss examples. As a result, a large fraction of gradient updates is spent on ``easy'' regions, while errors concentrated in a small but critical subset (often over-represented in target benchmarks) remain insufficiently optimized.

\textbf{(4) Conditional shift in outputs.} Finally, the conditional distribution can differ, i.e., $P_0(y\mid x)\neq P^*(y\mid x)$, which in LLM post-training often appears as differences in solution style, chain-of-thought structure, or answer formatting. Such mismatches can lead the model to learn a behavior prior that is misaligned with the target tasks.

\paragraph{How failure-driven retrieval-synthesis mitigates distribution mismatch.}
Our second-stage data construction explicitly targets these issues by inducing a new training distribution $P_{\mathrm{syn}}$ from model failures observed on a target set $Q$ (ideally, $Q\approx P^*$). Failure-driven resampling shifts training focus toward regions where the current model performs poorly under the target distribution, mitigating covariate shift and the underweighting of high-loss regions. Moreover, embedding-based retrieval defines a kernel
\begin{equation}
\label{eq:retrieval_kernel_discuss}
K(d\mid x)\propto \exp\!\big(\langle \phi(x),\phi(d)\rangle/\tau\big),
\end{equation}
which maps failure queries to relevant external documents. By synthesizing new training instances conditioned on the retrieved documents, we effectively expand the support of the training data, addressing support mismatch by assigning non-zero probability to regions that were absent in $P_0$. Finally, the synthesis prompts and verification filters allow us to better align output formats and reasoning traces with target requirements, partially reducing conditional shift.

\subsection{Failure-driven Post-training Provides Better Gradient Estimation}
\label{app:grad_alignment}


\begin{examplebox}
\begin{theorem}[Failure-driven training minimizes the expected loss.]
Let $\nabla \gL^*(\theta)$ be the ideal target graident, $\nabla \gL_{1}(\theta)$ be the stage-2 gradient under the synthetic
distribution $P_{1}$, and $\nabla \gL_0(\theta)$ be the gradient under the original
distribution $P_0$. Assuming the gradient is normalized such that $\|\nabla \gL_1(\theta)\|=\|\nabla \gL_0(\theta)\|$ and $Q\approx P*$, the failure-driven construction of $P_1$ makes the gradients better aligned with the target gradient, i.e.,
\begin{equation}
\label{eq:grad_sim}
\langle \nabla \gL^*(\theta), \nabla\gL_1(\theta)\rangle
\ge
\langle \nabla \gL^*(\theta), \nabla \gL_0(\theta)\rangle,
\end{equation}
Then, for sufficiently small learning rate $\eta$, one stage-2 optimization step using $P_{1}$ yields no larger target risk than using $P_0$:
\begin{equation}
\gL^*(\theta-\eta \nabla \gL_1(\theta)) \le \gL^*(\theta-\eta \nabla \gL_0(\theta)).
\end{equation}
\end{theorem}
\end{examplebox}

\textbf{Proof}: 
We first prove~\cref{eq:grad_sim} as a lemma:

\begin{examplebox}
\begin{lemma}[Conditions for Failure-Driven Gradient Alignment]
\label{lemma:kernel_conditions}
The failure-driven gradient $\nabla \gL_1$ achieves better alignment with the target gradient $\nabla \gL^*$ than the vanilla gradient $\nabla \gL_0$, i.e.,
\begin{equation}
    \langle \nabla \gL^*(\theta), \nabla\gL_1(\theta)\rangle \ge \langle \nabla \gL^*(\theta), \nabla \gL_0(\theta)\rangle,
\end{equation}
provided that:
1) The evaluation distribution $Q$ sufficiently covers the target distribution and mimics $P^*$ around the failure region; 2) The synthesis kernel $G(y'|d)$ correctly mimic $P^*(y\mid x)$ and 3) the retrieval kernel $K(d\mid x)$ correctly obtain samples similar to $x$. 
\end{lemma}
\end{examplebox}

We partition the data space based on the performance of the current model $\theta$:
\begin{itemize}
    \item \textbf{Success Region ($\mathcal{S}$)}: $\mathcal{S} = \{x \mid w_{\theta}(x) = 0\}$. The model predicts correctly and the loss is negligible.
    \item \textbf{Failure Region ($\mathcal{F}$)}: $\mathcal{F} = \{x \mid w_{\theta}(x) = 1\}$. The model fails; loss is high.
\end{itemize}
Let $y^*(x)$ denote the ground truth label (from $P^*$) for input $x$. We assume that in the success region $\mathcal{S}$, the model is well-converged. The expected gradient is dominated by zero-mean noise:
\begin{equation}
    \mathbb{E}_{x \sim P(\cdot|\mathcal{S})} [\nabla \ell_\theta(x, y^*(x))] \approx \mathbf{0}.
\end{equation}
Consequently, the target gradient is determined by the failure region. Let $\mathbf{v}^*$ be the average gradient direction required to fix the failures:
\begin{equation}
    \nabla \gL^*(\theta) \approx P^*(\mathcal{F}) \cdot \underbrace{\mathbb{E}_{x \sim P^*(\cdot|\mathcal{F})} [\nabla \ell_\theta(x, y^*(x))]}_{\triangleq \mathbf{v}^*}.
\end{equation}
Let the synthetic gradient be $\mathbf{v}_{\mathrm{syn}} = \mathbb{E}_{P_{\mathrm{syn}}}[\ell_\theta(x,y)]$, and assume a similarity factor $\alpha > 0$:
    \begin{equation}
        \label{eq:kernel_validity}
        \langle \mathbf{v}_{\mathrm{syn}}, \mathbf{v}^* \rangle \ge \alpha \|\mathbf{v}^*\|^2.
    \end{equation}
Then we are able to compare the alignment of the vanilla gradient $\nabla \gL_0$ and the failure-driven gradient $\nabla \gL_1$ with the target gradient $\nabla \gL^*$. The original training distribution $P_0$ is dominated by easy samples. Let $\epsilon = P_0(\mathcal{F})$ be the proportion of failure cases in the original dataset, we have
\begin{equation}
 \nabla \gL_0(\theta) = (1-\epsilon)\mathbb{E}_{\gS}[\ell_\theta] + \epsilon \mathbb{E}_{\mathcal{F}}[\ell_\theta]\approx \mathbf{0} + \epsilon \mathbf{v}^*.
\end{equation}
The inner product with the target gradient is:
\begin{equation}
\label{eq:align_g0}
    \langle \nabla \gL^*,  \nabla \gL_0 \rangle \approx \langle P^*(\mathcal{F})\mathbf{v}^*, \epsilon \mathbf{v}^* \rangle = \epsilon P^*(\mathcal{F}) \|\mathbf{v}^*\|^2.
\end{equation}
Then, recall $P_1 = \lambda P_0 + (1-\lambda) P_{\mathrm{syn}}$.
The synthetic distribution $P_{\mathrm{syn}}$ concentrates mass on $\mathcal{F}$ and generates gradients $\mathbf{v}_{\mathrm{syn}}$.
\begin{equation}
    \nabla \gL_1(\theta) = \lambda \nabla\gL_0(\theta) + (1-\lambda) \mathbf{v}_{\mathrm{syn}}.
\end{equation}
The inner product with the target gradient is:
\begin{equation}
    \langle \nabla \gL^*, \nabla \gL_1 \rangle = \lambda \langle \nabla \gL^*, \nabla\gL_0 \rangle + (1-\lambda) \langle \nabla \gL^*, \mathbf{v}_{\mathrm{syn}} \rangle.
\end{equation}
Substituting $\nabla \gL^* \approx P^*(\mathcal{F})\mathbf{v}^*$:
\begin{equation}
\label{eq:align_g1}
    \langle \nabla \gL^*, \mathbf{v}_{\mathrm{syn}} \rangle \approx P^*(\mathcal{F}) \langle \mathbf{v}^*, \mathbf{v}_{\mathrm{syn}} \rangle \ge \alpha P^*(\mathcal{F}) \|\mathbf{v}^*\|^2.
\end{equation}
From Eq.~\eqref{eq:align_g0} and Eq.~\eqref{eq:align_g1}, it is realistic to assume $\alpha \ge \epsilon$, then
\begin{equation}
    \alpha P^*(\mathcal{F}) \|\mathbf{v}^*\|^2 \ge \epsilon P^*(\mathcal{F}) \|\mathbf{v}^*\|^2
\end{equation}
which implies that 
\begin{equation}
\langle \nabla \gL^*(\theta), \nabla\gL_1(\theta)\rangle
\ge
\langle \nabla \gL^*(\theta), \nabla \gL_0(\theta)\rangle,
\end{equation}

\textbf{Remark.} The inequality holds when $\alpha \ge \epsilon \approx P_0(\mathcal{F})$, where $\alpha$ is the similarity factor between synthetic gradient and true gradient (typically high approximating 1), and $P_0(\mathcal{F})$ is the error rate on the training set (typically very low (e.g., $<0.1$) as the model fits the training data). $\alpha$ is determined by cetrain properties of $Q$, $K$ and $G$, which we discussed as,
\begin{enumerate}
    \item \textbf{$Q$ has to approximate $P^*$}. When $Q \approx P^*$, the evaluation distribution $Q$ effectively covers the target failure region, i.e., $Q_{\theta}(\mathcal{F}) \approx 1$. 
    \item \textbf{$K$ has to correctly retrieve the relevant documents}. So the retrieved $z$ are close to $x$ around the failure region. 
    \item \textbf{$G(x,y\mid d)$ jointly need to be close to $P^*(x,y)$} and $G(y'|d,x)$ produces synthetic labels $y'$ that provide a descent direction aligned with the ground truth, i.e. $y^\prime (x) \approx y^*(x)$. If retrieval is irrelevant or synthesis hallucinates (invalid $K, G$), then $\alpha \le 0$ in~\cref{eq:kernel_validity}, and the method fails.
\end{enumerate}

\begin{examplebox}
\begin{lemma}[Gradient alignment yields larger one-step decrease of the target risk]
Let $\gL^*(\theta)=\mathbb{E}_{(x,y)\sim P^*}[\ell_\theta(x,y)]$ be the target risk.
Assume $\gL^*$ is $L$-smooth, i.e., for all $\theta,\theta'$,
\begin{equation}
\|\nabla \gL^*(\theta)-\nabla \gL^*(\theta')\|\le L\|\theta-\theta'\|.
\end{equation}
Consider two update directions $g_0$ and $g_1$ at the same parameter $\theta$, and define the updates
\begin{equation}
\theta_i^+ = \theta - \eta g_i,\quad i\in\{0,1\}.
\end{equation}
If the step size satisfies $\eta\le 1/L$ and the two directions have the same norm
$\|g_0\|=\|g_1\|$, then
\begin{equation}
\langle \nabla \gL^*(\theta), g_0\rangle \le\langle \nabla \gL^*(\theta), g_1\rangle
 \Longrightarrow 
\gL^*(\theta_1^+) \leq \gL^*(\theta_0^+).
\end{equation}
In particular, among update directions of the same norm, a larger inner product with the target gradient
(i.e., better alignment) guarantees a smaller target risk after one step.
\end{lemma}
\end{examplebox}
\textbf{Proof}. Consider a single gradient-based update
\begin{equation}
\label{eq:gd_update}
\theta^+ = \theta - \eta g,
\end{equation}
where $g$ is the update direction. Using a first-order Taylor approximation of $L^*$ around $\theta$ yields
\begin{equation}
\label{eq:taylor_first_order}
\gL^*(\theta^+) \approx \gL^*(\theta) + \left\langle \nabla \gL^*(\theta), \theta^+-\theta \right\rangle
= \gL^*(\theta) - \eta \left\langle \nabla \gL^*(\theta), g \right\rangle.
\end{equation}
Eq.~\eqref{eq:taylor_first_order} shows that, for a fixed step size $\eta$,
the target risk decreases more in one step when the inner product
$\langle \nabla L^*(\theta), g\rangle$ is larger.

Assume that $\gL^*$ is $L$-smooth (i.e., its gradient is $L$-Lipschitz). Then the descent lemma gives:
\begin{equation}
\label{eq:descent_lemma}
\gL^*(\theta^+) \le
\gL^*(\theta)
- \eta \left\langle \nabla \gL^*(\theta), g \right\rangle
+ \frac{L\eta^2}{2}\|g\|^2.
\end{equation}
Therefore, for the same step size $\eta$, if $\langle \nabla \gL^*(\theta), g_0\rangle \leq\langle \nabla \gL^*(\theta), g_1\rangle$ makes the update reduce
the target risk more effectively, i.e.,
\begin{equation}
\gL^*(\theta_1^+) \leq \gL^*(\theta_0^+).
\end{equation}
For stage-2 SFT we typically take
\begin{equation}
\label{eq:g_is_grad_Lsyn}
g_0 = \nabla \gL_{0}(\theta),\quad g_1 = \nabla \gL_{1}(\theta).
\end{equation}
Substituting ends the proof. 

\subsection{Extension to RL: Theory}
\label{apdx: ext_RL}

Our failure-driven post-training framework can be naturally extends to reinforcement learning with verifiable rewards (RLVR). We first show that RL (with KL regularization) can be viewed as distribution matching
towards a reward-tilted target distribution, and that failure-driven RL further improves the match to the gold distribution $P^*$ by correcting the mismatch.

\paragraph{RL with KL regularization induces a reward-tilted target distribution.}
Let $\pi_\theta(y\mid x)$ denote the current policy and $\pi_0(y\mid x)$ be a reference policy (the stage-1 SFT model),
which serves as a strong proposal distribution with broad support.
Consider the following KL-regularized RL objective~\citep{DBLP:conf/icml/0011C00LZ23}:
\begin{equation}
\max_{\pi(\cdot\mid x)}
\mathbb{E}_{y\sim \pi(\cdot\mid x)}\big[ A(x,y) \big]
-\frac{1}{\beta}\mathrm{KL}\!\left(\pi(\cdot\mid x)\,\|\,\pi_0(\cdot\mid x)\right),
\label{eq:kl_rl_objective}
\end{equation}
where $A(x,y)$ is the advantage of some verifiable reward (e.g., correctness/format/length), and $\beta>0$ controls the strength of
the KL penalty. For any fixed $x$, the maximizer of~\cref{eq:kl_rl_objective} admits a closed form:
\begin{equation}
\pi^*_\beta(y\mid x)
=
\frac{\pi_0(y\mid x)\exp\big(\beta A(x,y)\big)}
{\sum_{y'}\pi_0(y'\mid x)\exp\big(\beta A(x,y')\big)}
\propto
\pi_0(y\mid x)\exp\!\big(\beta A(x,y)\big).
\label{eq:reward_tilted_optimal_policy}
\end{equation}
\Cref{eq:reward_tilted_optimal_policy} shows that KL-regularized RL performs an exponential tilting
of the proposal $\pi_0$ by the advantage, thereby inducing an implicit target conditional distribution.

Moreover, ~\cref{eq:kl_rl_objective} is equivalent to a KL-based distribution matching problem:
\begin{equation}
\pi^*_\beta(\cdot\mid x)
=
\arg\min_{\pi(\cdot\mid x)}
\mathrm{KL}\!\left(\pi(\cdot\mid x)\,\|\,\pi^*_\beta(\cdot\mid x)\right).
\label{eq:kl_matching_form}
\end{equation}
Therefore, a second-stage RL procedure can be interpreted as pushing $\pi_\theta(\cdot\mid x)$ towards
$\pi^*_\beta(\cdot\mid x)$, i.e., matching a target distribution $P^*(y\mid x)$.

\paragraph{Failure-driven RL improves the fit to $P^*$ by correcting the prompt marginal.}
In practice, the overall goal is to reduce the target risk under the unknown gold distribution $P^*(x,y)$.
Even if RL improves the conditional distribution $P(y\mid x)$, a mismatch on the prompt marginal
can still lead to suboptimal optimization signal. Concretely, standard RL typically samples prompts from a
surrogate prompt distribution (e.g., the stage-1 prompt marginal $P_0(x)$), while the true target prompt
marginal under $P^*$ may emphasize different regions (especially hard STEM problems).

Our failure-driven data construction defines a new prompt distribution
\begin{equation}
P_1(x) \;=\; \lambda P_0(x) + (1-\lambda)P_{\mathrm{syn}}(x),
\label{eq:p1_prompt_marginal}
\end{equation}
As discussed previously, $P_1(x)$ better mimic the real distribution $P^*(x)$, thus provides a better estimation to the target distribution. As a result, RL updates are computed using samples that are more representative of the hard regions that dominate the target risk, yielding a better approximation to the ideal distribution-matching objective. Under $P_1(x)$, the KL-regularized RL objective becomes
\begin{equation}
\max_\theta
\;\;
\mathbb{E}_{x\sim P_1(x),\,y\sim \pi_\theta(\cdot\mid x)}\!\big[ A(x,y) \big]
\;-\;
\frac{1}{\beta}\,\mathbb{E}_{x\sim P_1(x)}\!\left[\mathrm{KL}\!\left(\pi_\theta(\cdot\mid x)\,\|\,\pi_0(\cdot\mid x)\right)\right].
\label{eq:rl_under_p1}
\end{equation}
Thus, analogous to~\cref{eq:reward_tilted_optimal_policy}, ~\cref{eq:rl_under_p1} corresponds to matching the joint distribution
\begin{equation}
P^{*}_{\beta,1}(x,y)
\propto P_1(x)P_0(y\mid x)\exp\big(\beta A(x,y)\big),
\label{eq:joint_target_under_p1}
\end{equation}
where $P_1(x)$ is the surrogate prompt distribution and $P^\prime(y\mid x)\propto P_0(y\mid x)\exp\big(\beta A(x,y)\big)$ is the surrogate Response distribution. Overall $P^{*}_{\beta,1}(x,y)$ provides better estimated risk over the gold distribution $P^*(x,y)$ when $P_1(x)$ is a better surrogate of the target prompt marginal and $P^\prime(y\mid x)$ is a good Response distribution. . The proof follows almost the same as~\cref{app:grad_alignment}, thus we omit it here.

\end{document}